\pdfoutput=1

\documentclass{article}

\usepackage[final]{neurips_2024}

\usepackage[utf8]{inputenc} 
\usepackage[T1]{fontenc}    
\usepackage{hyperref}       
\usepackage{url}            
\usepackage{booktabs}       
\usepackage{amsfonts}       
\usepackage{nicefrac}       
\usepackage{microtype}      
\usepackage{xcolor}         

\usepackage{amsmath} 
\usepackage{soul}
\usepackage{listings}
\usepackage{wrapfig}
\usepackage{multirow}
\usepackage{graphicx}
\usepackage{caption}
\usepackage[normalem]{ulem}
\useunder{\uline}{\ul}{}
\usepackage{bbm}

\usepackage{colortbl}
\definecolor{Gray}{gray}{0.9}

\usepackage[breakable, most]{tcolorbox}
\definecolor{myframecolor}{HTML}{8babbe}
\definecolor{mycolback}{HTML}{e6e6e6}

\newcounter{definition}
\renewcommand{\thedefinition}{\arabic{definition}}
\newcommand{\mydefinition}[1]{%
  \refstepcounter{definition}
  \textbf{Definition \thedefinition}\label{#1}%
}

\title{Beyond Accuracy: Ensuring Correct Predictions With Correct Rationales}

\author{%
  Tang Li \quad \quad Mengmeng Ma \quad \quad Xi Peng \\
  \\
  DeepREAL Lab: \url{https://deep-real.github.io} \\
  Department of Computer \& Information Science, University of Delaware \\
  \texttt{\{tangli, mengma, xipeng\}@udel.edu} \\
}

\begin{document}

\maketitle

\begin{abstract}
Large pretrained foundation models demonstrate exceptional performance and, in some high-stakes applications, even surpass human experts. However, most of these models are currently evaluated primarily on prediction accuracy, overlooking the validity of the rationales behind their accurate predictions. For the safe deployment of foundation models, there is a pressing need to ensure {\it double-correct predictions}, {\it i.e.}, correct prediction backed by correct rationales. To achieve this, we propose a two-phase scheme: First, we curate a new dataset that offers structured rationales for visual recognition tasks. Second, we propose a rationale-informed optimization method to guide the model in disentangling and localizing visual evidence for each rationale, without requiring manual annotations. Extensive experiments and ablation studies demonstrate that our model outperforms state-of-the-art models by up to 10.1\% in prediction accuracy across a wide range of tasks. Furthermore, our method significantly improves the model's rationale correctness, improving localization by 7.5\% and disentanglement by 36.5\%. Our dataset, source code, and pretrained weights: \href{https://github.com/deep-real/DCP}{https://github.com/deep-real/DCP}
\end{abstract} 
\section{Introduction}
\label{sec:introduction}
\begin{wrapfigure}{r}{0.45\textwidth}
\vspace{-18pt}
  \begin{center}
  \includegraphics[width=0.45\textwidth]{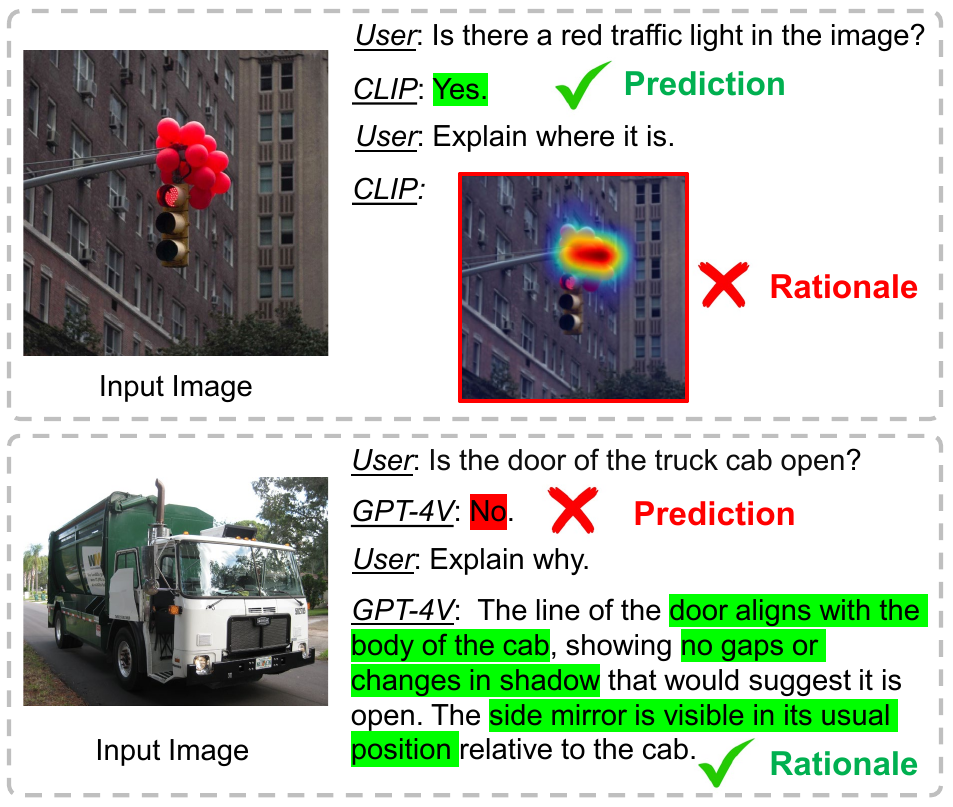}
  \end{center}
  \vspace{-8pt}
  \caption{
  Unsafe prediction examples.
  \textbf{Correct prediction, incorrect rationale}: CLIP identifies a red light, but wrongly based on red balloons.
  \textbf{Incorrect prediction, correct rationale}: GPT-4V incorrectly predicts a closed door, yet based on plausible visual evidence.
  }
  \label{fig:title}
\vspace{-15pt}
\end{wrapfigure}
Large foundation models, such as CLIP~\cite{radford2021learning} and GPT-4V~\cite{openai2023gpt4}, exhibit exceptional performance or even surpass human experts in some high-stakes applications, such as medical diagnosis~\cite{nori2023capabilities} and autonomous driving~\cite{liang2022effective, liang2023visual}.
However, most of these models are currently evaluated primarily on prediction accuracy, overlooking a critical aspect for ensuring safety, {\it i.e.}, the validity of the reasons behind their accurate predictions.
Understanding the {\it rationales} - the ``how'' and ``why'' behind model predictions - is crucial for developing safe predictions.
Fig.~\ref{fig:title} shows typical examples of unsafe predictions:
CLIP might predict accurately yet based on wrong rationales, whereas GPT-4V might make wrong predictions based on rationales that are plausible to humans.
To build trust in real-world deployment, a natural question arises: 
{\it Can models make double-correct predictions, i.e., correct predictions backed by correct rationales?}

Correct rationales generally align with how humans would reason about the same decision and are based on valid {\it visual evidence}~\cite{teach1981analysis, ribeiro2016should, rudin2019stop}.
There are existing attempts to provide rationales for machine learning models' predictions.
They either explicitly force the models to make decisions based on human-understandable concepts by introducing bottleneck layers~\cite{menon2022visual, yuksekgonul2022post}, or implicitly inject commonsense knowledge into models by contrastive learning between similar yet distinct textual concepts~\cite{yuksekgonul2022and, ye2023improving}.
However, none of them ensures {\it double-correct predictions}.
Observations from our previous research~\cite{li2025deal} and recent studies in the field~\cite{margeloiu2021concept, tong2024eyes} reveal that these models might provide {\it incorrect rationales}, as they fail to base the rationales on valid {\it visual evidence}.

To this end, we develop \textit{double-correct predictions} by focusing on two foundational aspects: 

{\bf i) ``What'' are the correct rationales? Structured rationale acquisition.}
Existing vision datasets typically provide ground truth labels of predictions, whereas missing the rationales behind these decisions~\cite{fei2006one, wah2011caltech}.
To fill this gap, we curate a new dataset that offers over 4,000 unique {\it textual rationales} designed for predicting the 1,000 categories in {\it ImageNet}~\cite{deng2009imagenet}, structured in a tree format.
This design differs from existing knowledge graphs~\cite{speer2017conceptnet, miller1995wordnet, krishna2017visual}, which either provide irrelevant knowledge for the vision task or are too coarse-grained, providing insufficient information.
Our rationale dataset is tailored to capture the detailed reasoning processes for visual recognition.

{\bf ii) ``Where'' are the correct rationales? Rationale-informed optimization.} 
The other challenge in developing double-correct predictions is the absence of pixel-wise annotations for rationales' {\it visual evidence}.
Although some datasets provide segmentation masks of object parts~\cite{wah2011caltech, tschandl2018ham10000}, they lack sufficient rationale coverage and are limited to small-scale use cases~\cite{roscher2020explainable}.
To address this issue, we propose a rationale-informed optimization method to guide the model in disentangling and localizing the visual evidence of rationales, without requiring manual annotations.
Our method can be integrated into the existing model training process without architectural changes and extra parameters.

We evaluate the proposed method on a wide range of benchmark datasets and tasks.
For prediction correctness, our model outperforms state-of-the-art models in zero-shot, linear probe, and fine-tuning settings by 2.6\%, 2.0\%, and 10.1\%.
For rationale correctness, the empirical results exhibit that our model significantly improves ground truth rationale localization and rationale disentanglability by 7.5\% and 36.5\%.
Furthermore, the extensive qualitative results and ablation studies demonstrate the effectiveness of the proposed method.

Our contribution includes:
{\bf 1)} We curate a new structured rationale dataset.
{\bf 2)} A faithful explanation method tailored for explaining CLIP-ViT predictions.
{\bf 3)} A principled optimization method that seamlessly integrates structured rationale information to develop {\it double-correct predictions}.
{\bf 4)} Empirical results in a wide range of benchmark datasets and tasks including image classification and retrieval demonstrate the superior prediction and rationale correctness of our model.
\section{Problem Formulation}
\label{sec:problem}
In this section, we first formally define {\it rationales}, then provide the mathematical formulation of the {\it double-correct prediction} problem. 

\mydefinition{def:rationale}
{\it (Rationales) Given a category $y$, rationales are a set of $K$ underlying abstract notions $\{r_k^y\}_{k=1}^K$ and relations that capture the reasoning process leading to the recognition of $y$.}

In the real world, rationales can be represented through textual descriptions~\cite{barsalou2008grounded, siskind1994grounding}.
For example, when recognizing a specific breed of dog in an image, the rationales could be a set of concepts such as the shape of the ears, the color of the fur, and the size of the dog.
Mathematically, given a textual rationale $r$, we assume the existence of a ground truth labeling function $V(x, r)$ that can provide the pixel-wise annotations of {\it visual evidence} corresponding to $r$ on an input $x$.

\mydefinition{def:double-correct}
{\it (Double-Correct Predictions) A correct prediction is double-correct when it is backed by correct rationales that are based on valid visual evidence.}

Denote $(x, y) \sim P(X,Y)$ as a data point sampled from the training distribution $P(X,Y)$, $g(\cdot)$ as an explanation method that attributes the prediction of text $r$ to a group of pixels in input $x$ depending on model $f$, $\ell (\cdot)$ as the task-specific loss function, and $\mathcal{F}$ as a function class that is model-agnostic for the prediction task.
To ensure the model $f$ makes {\it double-correct prediction}, we propose to solve the following constrained optimization problem:
\begin{equation}
\mathop{\mathrm{min}}\limits_{f\in \mathcal{F}} \mathcal{R}(f) : = \mathbb{E}_{(x,y)\sim P(X,Y)} [\ell(f(x), y)]
\quad \; \mathrm{s.t.} \;\;  g(x,r;f) = V(x,r), \;\; \forall r\in \{r_k^y\}_{k=1}^K .
\label{eq:dual_correct_prediction}
\end{equation}
The problem in Eq.~\ref{eq:dual_correct_prediction} is challenging to solve, since we neither have access to the rationales $\{r_k^y\}_{k=1}^K$, nor to the ground truth labeling functions $V(\cdot)$.
There are existing attempts that employ domain experts to manually collect textual descriptions of rationales~\cite{tschandl2018ham10000, daneshjou2022skincon}, or pixel-wise annotations of object parts on the image~\cite{wah2011caltech}.
However, these approaches are often limited to small-scale datasets, and impractical in large-scale settings due to the high cost of fine-grained annotations~\cite{wang2020self, roscher2020explainable}.

\section{Double-Correct Predictions}
\label{sec:method}
To bridge the gaps, in Sec.~\ref{subsec:dataset} we present how to acquire rationales $\{r_k^y\}_{k=1}^K$, in Sec.~\ref{subsec:decompose} we propose a new explanation method $g(\cdot)$, and in Sec.~\ref{subsec:rationale-informed} we develop {\it double-correct predictions} without $V(\cdot)$.

\subsection{Structured Rationale Dataset}
\label{subsec:dataset}

\setlength{\intextsep}{2pt}%
\setlength{\columnsep}{10pt}%
\begin{wrapfigure}{r}{0.45\textwidth}
\vspace{-7pt}
  \begin{center}
  \includegraphics[width=0.45\textwidth]{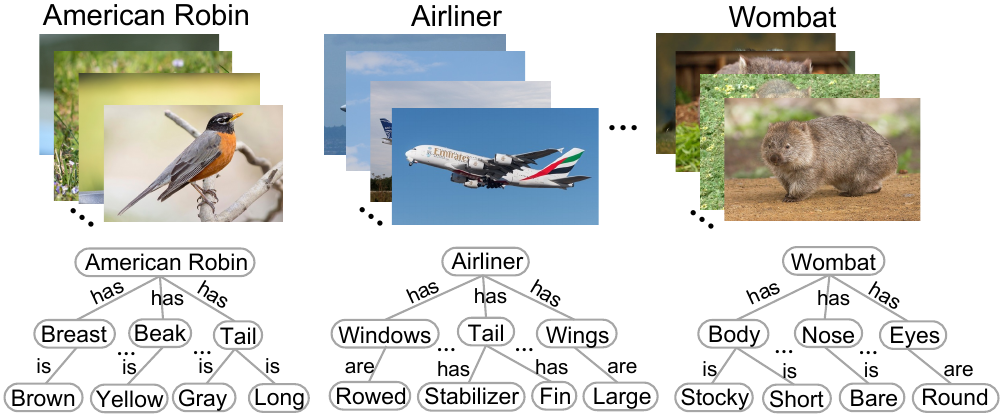}
  \end{center}
  \vspace{-8pt}
  \caption{
  Our structured rationales capture the major attributes and their sub-attributes that lead to the recognition of objects.
  Our dataset offers over 4,000 unique rationales covering all 1,000 categories from {\it ImageNet}~\cite{deng2009imagenet}.
  }
  \label{fig:ontology}
\end{wrapfigure}
In this section, we curate a new rationale dataset to offer $\{r_k^y\}_{k=1}^K$ in Eq.~\ref{eq:dual_correct_prediction}.
According to Def.~\ref{def:rationale}, rationales are structured human knowledge.
Therefore, {\it ontologies} that encapsulate complex, interconnected information while maintaining semantic relationships between entities~\cite{wang2017knowledge, ji2021survey}, present a proper tool to represent rationales.
The benefits are bi-directional:
{\bf i)} in the human-to-machine direction, it offers a standardized, machine-readable format;
{\bf ii)} in the machine-to-human direction, ontology structure mirroring how humans organize and retrieve information to explain the model's decision-making process.

{\bf Acquire structured rationales:}
Different from existing works that are limited to small-scale manual annotation, we generate our rationale dataset in a scalable manner.
Specifically, we utilize Large Language Models (LLMs) like GPT-4~\cite{openai2023gpt4} to extract the structured rationales.
Existing studies prove that GPT-4 has expert-level expertise in commonsense~\cite{bubeck2023sparks} and domain knowledge~\cite{liu2023holistic}.
However, we find that directly querying LLMs would yield inconsistent tree structures that can hardly be used by machine learning models.
To address this issue, we provide a series of exemplary structured rationales before the query, employing in-context learning~\cite{brown2020language} to extract {\it standardized} rationales in a \texttt{.JSON} format.
See Appendix~\ref{appen_subsec:prompts} for our full prompt and rationale examples.

{\bf Rationale dataset statistics:}
Our dataset covers all 1,000 categories in the {\it ImageNet}~\cite{deng2009imagenet}.
For each category, we generate an ontology tree with a maximum height of two.
As illustrated in Fig.~\ref{fig:ontology}, the root node is the category, the children of the root are the attributes, and the leaves are the sub-attributes.
The edges represent the relationships between nodes.
Combining attributes and sub-attributes, our dataset contains over 4,000 unique rationales.
Our rationale ontology trees capture the reasoning processes leading to the recognition of the corresponding root categories.

{\bf Can we trust the rationales extracted from GPT-4?}
Although there are plenty of works showing GPT-4's remarkable capabilities~\cite{bubeck2023sparks, liu2023holistic}, it still could suffer from {\it hallucinations}~\cite{mcintosh2023culturally, chelli2024hallucination}.
However, evaluations on the generation quality are largely missing from existing works that generate data from LLMs~\cite{menon2022visual, yang2023language, qin2022medical}.
To fill this gap and ensure the quality of our rationale data, we conduct comprehensive human and machine evaluations.
As detailed in Sec.~\ref{subsec:rationale_quality}, on a 5-point Likert scale across three metrics, 964 out of 1,000 categories are scored as having high-quality rationales ($\ge$4.0).

In contrast to existing Knowledge Graphs~\cite{speer2017conceptnet, miller1995wordnet, krishna2017visual} that either offer knowledge {\it unrelated} to the visual prediction task, or are too coarse-grained that provide {\it insufficient} information, our structured rationales are tailored for visual recognition tasks in a fine-grained attribute level.
Furthermore, our dataset can expand to accommodate new rationales, providing flexibility to dealing with evolving datasets where more data becomes available.
For example, our rationale ontologies can be seamlessly integrated following the {\it ImageNet}~\cite{deng2009imagenet} category ontology derived from WordNet~\cite{miller1995wordnet}.


\subsection{Faithful Explanation Method}
\label{subsec:decompose}
In this section, we develop a new explanation method to implement $g(\cdot)$ in Eq.~\ref{eq:dual_correct_prediction}.
To incorporate both image and text inputs, we instantiate the model $f$ using the CLIP-ViT architectures~\cite{dosovitskiy2020image} because of their proven capability~\cite{radford2021learning, zhou2022conditional}.
Existing methods for explaining the ViT model either directly use the attention maps as explanations~\cite{abnar2020quantifying}, or weigh them using gradients~\cite{chefer2021transformer,selvaraju2017grad}.
However, these methods might be {\it unfaithful} to the ViT predictions.
This is because the computation of each ViT prediction involves queries, keys, and values, whereas the attention maps only capture the inner products of queries and keys, ignoring information in values that also affect predictions~\cite{liu2022rethinking, chefer2021transformer}.
Therefore, explanations based on attention maps might not fully reflect the reasons behind ViT predictions.

{\bf Decompose ViT outputs:}
Recent works~\cite{elhage2021mathematical, gandelsman2023interpreting} prove that, for ViT models, the image embeddings can be decomposed into the contributions of each token within each attention head.
Let $\phi$ and $\theta$ parameterize the image- and text-encoder of the CLIP-ViT model, $P$ is the projection matrix, $L$, $M$, $N$ are the numbers of layers, heads, and image tokens, $a_i^{l,m}$ is the output of the $m$-th attention head in layer $l$ for the $i$-th image token, then the embedding of image $I$ can be decomposed as:
\begin{equation}
e_I = f_\phi(I) = P\mathrm {ViT}(I) = {\textstyle \sum_{l=1}^{L}} {\textstyle \sum_{m=1}^{M}} {\textstyle \sum_{i=0}^{N}} Pa_i^{l,m} .
\label{eq:msa_decompose}
\end{equation}
By contracting along layers and heads, \cite{gandelsman2023interpreting} calculates the contribution of the $i$-th image token to the final image embedding using ${\textstyle \sum_{l=1}^{L}} {\textstyle \sum_{m=1}^{M}} Pa_i^{l,m}$.

\begin{figure}[t]
    \centering
    \begin{minipage}{0.45\textwidth}
        \centering
        \includegraphics[width=1.0\textwidth]{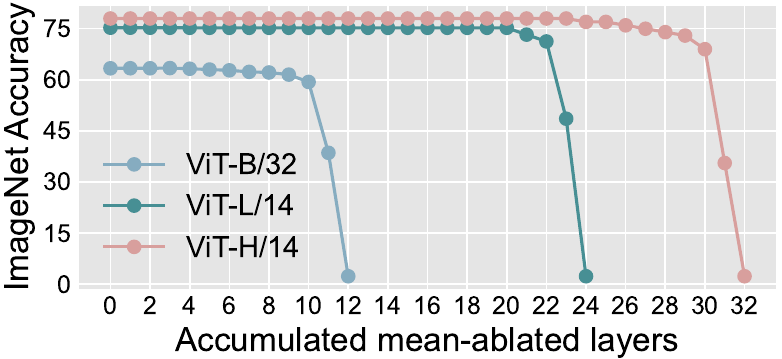}
        \caption{Multi-head Self Attention (MSA) accumulated mean-ablation study.
        Based on Eq.~\ref{eq:msa_decompose}, we replace the direct effects of MSAs up to a specific layer with their mean values calculated across the {\it ImageNet}~\cite{deng2009imagenet} validation set.
        Most of the performance gains can be attributed to the final layers of the ViT.
        }
        \label{fig:mean-ablation}
    \end{minipage}
    \hfill 
    \begin{minipage}{0.52\textwidth}
        \centering
        \captionof{table}{
        Weakly-supervised segmentation accuracy on {\it ImageNet-Seg}~\cite{guillaumin2014imagenet}.
        We threshold explanation heatmaps from CLIP-ViT-L-14 as segmentation masks.
        Our method outperforms existing explanation methods in segmentation accuracy, demonstrating the high faithfulness of our explanations.
        }
        \resizebox{1.0\columnwidth}{!}{%
\begin{tabular}{lccc}
\toprule[0.75pt]
Exp. Methods  & Pixel Acc. $\uparrow$ & mIoU $\uparrow$ & mAP $\uparrow$ \\ \toprule[0.5pt]
LRP~\cite{binder2016layer}           & 52.81                 & 33.57           & 54.37          \\
rollout~\cite{abnar2020quantifying}       & 60.63                 & 40.64           & 74.47          \\
row attention & 65.67                 & 43.83           & 76.05          \\
GradCAM~\cite{selvaraju2017grad}       & 70.27                 & 44.50           & 70.30          \\
Chefer et al.~\cite{chefer2021transformer} & 69.21                 & 47.47           & 78.29          \\
TextSpan~\cite{gandelsman2023interpreting}      & 75.21                 & 54.50           & 81.61          \\
\rowcolor{Gray}Ours          & \textbf{76.27}        & \textbf{58.04}  & \textbf{82.17} \\ \toprule[0.75pt]
\end{tabular}
}
\label{tab:imagenet-seg}
    \end{minipage}
\end{figure}

{\bf Faithful explanations weighted by mean-ablation results:}
As indicated by our mean-ablation results in Fig.~\ref{fig:mean-ablation}, the final layers contribute the most to the predictions, whereas the earlier layers have minimal impact.
Thus, noise from early layers could obscure key information by a naive summation across all layers as in \cite{gandelsman2023interpreting}.
To address this issue, we weigh each layer's contribution based on its importance, measured by the corresponding performance drop in the mean-ablation study.
Denote the performance drop of layer $l$ as $\Delta_l $, we calculate the contribution of the $i$-th image token by:
\begin{equation}
e_i  =  {\textstyle \sum_{l=1}^{L}} w^l {\textstyle \sum_{m=1}^{M}} Pa_i^{l,m}, \quad \mathrm{where} \;\; w^l = \frac{\Delta_l }{ {\textstyle \sum_{j=1}^{L}}\Delta_j} .
\end{equation}
Note that $e_i$ is projected onto the image-text embedding space by $P$. Thus, we can use $g(I,r) = \{ \langle e_i, f_\theta(r) \rangle\}_{i\in I}$ to calculate the explanations of rationale $r$ on an image $I$, {\it i.e.}, {\it visual evidence}.

Our method significantly improves the explanation accuracy, as shown in Tab.~\ref{tab:imagenet-seg}.
In contrast to attention-based explanations~\cite{abnar2020quantifying}, our method fully utilizes the information from queries, keys, and values that are used for ViT predictions.
Compared to gradient-weighted attention maps~\cite{chefer2021transformer,selvaraju2017grad}, our method cuts down the computational complexity from $O(n^2)$ to $O(n)$ over $n$ image tokens.

\subsection{Rationale-informed Optimization}
\label{subsec:rationale-informed}
In this section, we develop {\it double-correct predictions} by disentangling and localizing rationales without pixel-wise human annotations $V(\cdot)$ in Eq.~\ref{eq:dual_correct_prediction}.

{\bf Disentanglement via reconstruction:}
Drawing insights from our previous research~\cite{li2023dre, li2025deal}, we propose to contrast between explanation heatmaps of rationales to guide the model training in a {\it self-supervised} manner.
Specifically, we enforce the following two constraints: {\bf i)} the image embeddings for different rationales within the same category are disentangled, and {\bf ii)} the aggregated image embedding of all rationales within the same category aligns with the text embedding of the category.

Mathematically, the backbone objective is to learn a mapping function $f\in \mathcal{F}$ such that for each image-text pair $(I, T) \sim P(\mathbf{I}, \mathbf{T})$, the embeddings $f_\phi(I)$ and $f_\theta(T)$ are aligned in a shared space if they are a correct match, where $T$ is a text description of category $y$.
Let $\ell(\cdot)$ be the InfoNCE loss~\cite{oord2018representation}. $h(g(I,r)) =  {\textstyle \sum_{i}} e_i  \cdot \mathbbm{1} (g(I,r)_i > \tau)$ extracts the image embedding of a given rationale.
$\mathcal{D}(\cdot, \cdot)$ is a distance metric such as L2 distance.
$\tau$, $\epsilon$, and $\delta$ are thresholding hyperparameters.
For all $r, r' \in \{r_k^y\}_{k=1}^K$, we propose to develop {\it double-correct predictions} by optimizing:
\begin{equation}
\begin{aligned}
\min_{f \in \mathcal{F}} \; & \; \mathcal {R} (f) := \mathbb{E}_{(I,T) \sim P(\mathbf{I}, \mathbf{T})} [\ell(f_\phi(I), f_\theta(T))]  & & \lhd  {\small \textbf{Correct Predictions}} \\
\; \mathrm{s.t.} \, & \; \underbrace{\mathcal{D} (h(g(I,r)), h(g(I,r'))) \ge \epsilon,}_\textrm{\small Disentanglement} \;\; \underbrace{\mathcal{D} (\textstyle \sum_{r}^{} h(g(I,r)), f_\theta(y)) \le \delta.}_\textrm{\small Reconstruction} & & \lhd  {\small \textbf{Correct Rationales}}
\end{aligned}
\label{eq:disen_by_recon}
\end{equation}
Intuitively, the reconstruction term prevents the disentanglement from collapsing into trivial solutions, thereby ensuring localization.
Solving Eq.~\ref{eq:disen_by_recon} often leads to a non-convex problem, wherein methods such as stochastic gradient descent (SGD) cannot guarantee constraint satisfaction~\cite{robey2021model, qiao2023topology}.
To address this issue, we leverage Karush–Kuhn–Tucker (KKT) conditions~\cite{boyd2004convex, ma2024beyond} and introduce Lagrange multipliers $\lambda$ and $\gamma$ to convert the constrained problem into its unconstrained counterpart:
\begin{equation}
\begin{aligned}
\min_{f \in \mathcal{F}} \; \{\mathcal {R} (f) & := \mathbb{E}_{(I,T) \sim P(\mathbf{I}, \mathbf{T})} [\ell(f(I, T))] \\
& + \lambda \mathcal{D} (h(g(I,r)), h(g(I,r'))) \; + \; \gamma \mathcal{D} (\textstyle \sum_{r}^{} h(g(I,r)), f_\theta(y))\} .
\end{aligned}
\end{equation}

Our method has the following merits:
{\bf i)} In contrast to existing works that rely on expensive pixel-wise annotations to localize objects~\cite{gao2022pyramidclip, zeng2022multi}, the proposed rationale-informed optimization achieves a more fine-grained, attribute-level localization without manual annotations.
{\bf ii)} Our method can be integrated into vision-language model training without architectural changes and extra parameters.
\section{Experiments}
In this section, we first evaluate the quality of our curated rationale dataset in Sec.~\ref{subsec:rationale_quality}.
To best validate {\it double-correct predictions}, we then conduct a series of experiments to compare the proposed method with existing methods in Secs~\ref{subsec:benchmarks} -~\ref{subsec:exp_retrieval}.
The experimental results prove that our model achieves superior prediction and rationale correctness on a wide range of benchmark datasets and tasks.

\subsection{Evaluation of Rationale Quality}
\label{subsec:rationale_quality}


{\bf Metrics:}
We focus on three essential aspects of the rationale quality.
(1) {\it Factual Consistency}: whether the rationales are consistent with facts.
(2) {\it Comprehensiveness}: whether the rationales provide sufficient information necessary to predict the category.
(3) {\it Visual Disentanglement}: whether the rationales are visually disentanglable or non-overlap.
We rate them on a 5-point Likert scale scoring system, where higher scores indicate better performance.
For example, in Factual Consistency, score 5 means 100\% of the generated rationales are consistent with facts, score 4 means 75\%, score 3 means 50\%, score 2 means 25\%, and score 1 means completely wrong.

{\bf Evaluators:}
(1) {\it Human Evaluators}: We recruited four human evaluators, who are mostly graduate students.
They are asked to conduct assessments based on commonsense knowledge and perform Internet searches for validation.
On average, it takes them around one minute per sample.
(2) {\it Machine Evaluators}: The latest GPT-4o and GPT-4v models (date accessed: Aug. 6th, 2024).
For each evaluation, we perform three independent runs and calculate the average scores.
Note that expanding human evaluations to the entire dataset is not scalable.
To this end, we first prove the reliability of machine evaluations, then use it to automatically evaluation the entire dataset.

{\bf Human evaluations:}
We sample three independent groups of data from our rationale dataset, each consisting of 50 categories and their corresponding rationales. Specifically, categories were randomly selected from their superclasses: Animals (20), Objects \& Artifacts (15), Natural Scenes (5), Plants (5), and Human Activities (5).
This ensures that not only each superclass is represented but also that our results are robust~\cite{torralba2011unbiased}.
As shown in Tab.~\ref{tab:rationale_eval}, The dataset consistently achieves scores of 4.61 or higher on the average of evaluators for each metric, indicating that over 90.3\% of the rationales for each category are highly factual, comprehensive, and visually disentanglable.

{\bf Machine evaluations:}
Note that the scores of all three metrics are almost identical between machines and humans.
The Pearson Correlation coefficient of 0.82 reveals the strong positive correlation between machine and human evaluators.
Based on this observation, we further conduct machine evaluations on the entire dataset efficiently.
Our results indicate that 964 out of 1,000 categories have high-quality rationales ($\ge$4.0). See detailed results for the entire dataset in Appendix.~\ref{appen_subsec:rationale_eval}.

\begin{table}[t]
\centering
\caption{Evaluation results of rationale quality.
Both machine and human evaluators receive the same instructions about the metrics.
The scores for all three metrics are nearly identical between machine and human evaluators, indicating that over 90.3\% of our rationales are of high quality.
}
\vspace{3pt}
\resizebox{0.83\columnwidth}{!}{%
\begin{tabular}{lccc}
\toprule[0.75pt]
Evaluators   & Factual Consistency & Comprehensiveness & Visual Disentanglement \\ \toprule[0.5pt]
GPT-4o       & 4.89$\pm$0.05       & 4.55$\pm$0.06     & 4.66$\pm$0.06          \\
GPT-4v       & 4.92$\pm$0.03       & 4.67$\pm$0.05     & 4.70$\pm$0.02          \\
\rowcolor{Gray} 
Machine Avg. & 4.91                & 4.61              & 4.68                   \\ \toprule[0.5pt]
Human-A      & 4.85$\pm$0.11       & 4.64$\pm$0.19     & 4.42$\pm$0.15          \\
Human-B      & 4.97$\pm$0.02       & 4.77$\pm$0.02     & 4.20$\pm$0.11          \\
Human-C      & 4.78$\pm$0.04       & 4.60$\pm$0.11     & 4.78$\pm$0.10          \\
Human-D      & 4.81$\pm$0.08       & 4.65$\pm$0.05     & 4.77$\pm$0.07          \\
\rowcolor{Gray} 
Human Avg.   & 4.85                & 4.66              & 4.54                   \\ \toprule[0.75pt]
\end{tabular}
}
\label{tab:rationale_eval}
\end{table}

\subsection{Benchmark Datasets and Implementation Details}
\label{subsec:benchmarks}
{\bf Backbone model:}
Due to the computational cost of training large vision-language models (VLMs) from scratch, we focus on fine-tuning experiments.
Specifically, we fine-tune the ViT-B/32 variant of CLIP on the {\it ImageNet}~\cite{deng2009imagenet} dataset combined with our curated rationale dataset.
To maintain simple and interpretable rationales, the ontology graph for each category is limited to a maximum depth of two, allowing for the extraction of five to six independent concepts on average.

{\bf Baseline models:}
We compare our model with state-of-the-art VLMs that use ViT-B/32 as their vision encoders, including large-scale pretrained models (CLIP~\cite{radford2021learning}, DeCLIP~\cite{li2022supervision}), knowledge-augmented model (NegCLIP~\cite{yuksekgonul2022and}), and fine-grained alignment models (FILIP~\cite{yao2021filip}, PyramidCLIP~\cite{gao2022pyramidclip}).
For fair comparisons, we also compare our model with {\it ImageNet}~\cite{deng2009imagenet} fine-tuned models using the same CLIP initialization and augmented text descriptions as our model, including full model fine-tuning (-ft) and vision-encoder-only fine-tuning (-ft-vision).

{\bf Evaluation datasets:}
We validate the prediction correctness of the models on image classification and image-text retrieval tasks.
For image classification (zero-shot, linear probe), experiments are carried out on nine benchmark datasets, including {\it CUB}~\cite{wah2011caltech}, {\it Caltech101}~\cite{fei2004learning}, {\it OxfordPets}~\cite{parkhi2012cats}, {\it Food101}~\cite{bossard2014food}, {\it SUN397}~\cite{xiao2010sun}, {\it StanfordCars}~\cite{krause20133d}, {\it DTD}~\cite{cimpoi2014describing}, {\it CIFAR-10}~\cite{krizhevsky2009learning}, and {\it CIFAR-100}~\cite{krizhevsky2009learning}.
For retrieval, we conduct experiments on {\it Flickr30K}~\cite{young2014image} and {\it MSCOCO}~\cite{lin2014microsoft}.
To evaluate the correctness of rationales, we evaluate the models' rationale localizability on {\it CUB-Part}~\cite{saha2022improving} and {\it PartImageNet}~\cite{he2022partimagenet} that provide ground truth segmentation masks of object parts, {\it e.g.}, ``head'' and ``body''.
Furthermore, we evaluate the rationale disentanglability on the aforementioned nine benchmark datasets.
More details can be found in Appendix~\ref{appen_subsec:details}.

{\bf Implementation details:}
We follow the same architecture design as CLIP~\cite{radford2021learning} for ViT-B/32.
The input resolution of image encoder is 224$\times$224 and the maximum context length of text encoder is 77.
We train our model using an AdamW~\cite{loshchilov2018decoupled} optimizer and the cosine learning rate scheduler with a linear warmup.
Specifically, the learning rate linearly increases from 0 to the peak value within 10\% of the total steps, and then decreases with a cosine anneal strategy.
Our learning rate is set to 5e-7 and train the model for eight epochs.
More details can be found in Appendix~\ref{appen_subsec:details}.

\subsection{Evaluation Metrics}
\label{subsec:evaluation_metrics}
{\bf Prediction correctness:}
We use standard category {\it prediction accuracy} to evaluate the prediction correctness for zero-shot, linear probe, and fine-tuned settings.

{\bf Rationale correctness:}
We define two new metrics to measure rationale correctness.

i) {\it Rationale localizability}.
We evaluate the correctness of rationales using ground truth segmentation masks of object parts~\cite{saha2022improving, he2022partimagenet}.
Following the standard evaluation protocol~\cite{selva2017grad}, we threshold the rationale explanation heatmaps to segmentation masks and calculate a mean Intersection over Union (mIoU $\uparrow$) score with the ground truth masks across different object parts.
Specifically, the dynamic threshold $\tau =\mu +\sigma$, where $\mu$ and $\sigma$ are the mean and standard deviation of importance values of all pixels in a heatmap.
The pixel with an importance value larger than $\tau$ is set to 1, otherwise 0.

ii) {\it Rationale disentanglability}.
As shown in Fig.~\ref{fig:qualitative}, for the CLIP model~\cite{radford2021learning}, the visual evidence of different rationales is entangled.
Specifically, we treat the disentanglement between the visual evidence of different rationales as an important metric to evaluate whether the model can distinguish rationales.
Specifically, we treat rationale explanation heatmaps $\mathbf{m}$ and $\mathbf{m}'$ as vectors and calculate $1 - |\langle \mathbf{m}, \mathbf{m}' \rangle |$ as an intuitive measure of disentanglability, the higher metric value the better.


\begin{table}[t]
\centering
\caption
{Comparison of prediction accuracy (\%) on nine benchmark datasets.
Our results are on the average of three trials of experiments using different random seeds.
We highlight the {\bf best results} and the \underline{second best} results.
Surprisingly, different from most interpretability methods that compromise benchmark performance, our method also enhances prediction accuracy.
}
\resizebox{1.0\columnwidth}{!}{%
\begin{tabular}{clcccccccccc}
\toprule[1pt]
Metrics                                                                                             & Models         & C10                                & C100                               & CUB                                & CAL                                & PETS                               & F101                               & SUN                                & CARS                               & DTD                                & AVG                                \\ \toprule[0.75pt]
\multirow{8}{*}{\begin{tabular}[c]{@{}c@{}}Zero-shot\\ Accuracy (\%)\end{tabular}}                 & CLIP           & \textbf{91.3}                      & 65.1                               & {\ul 51.5}                         & 87.9                               & 87.0                               & \textbf{84.4}                      & 63.2                               & 59.4                               & 44.5                               & {\ul 70.5}                         \\
                                                                                                    & DeCLIP         & {\ul 91.2}                         & {\ul 66.4}                         & 51.2                               & \textbf{89.5}                      & 79.5                               & 74.6                               & 63.4                               & 50.6                               & 42.7                               & 67.7                               \\
                                                                                                    & NegCLIP        & 85.7                               & 60.9                               & 37.4                               & 81.0                               & 79.7                               & 71.1                               & 57.0                               & 45.4                               & 37.5                               & 61.7                               \\
                                                                                                    & FILIP          & 86.9                               & 65.5                               & 37.5                               & 91.9                               & {\ul 88.1}                         & 82.8                               & {\ul 69.1}                         & 55.4                               & 49.3                               & 69.6                               \\
                                                                                                    & PyramidCLIP    & 81.5                               & 53.7                               & 52.7                               & 81.7                               & 83.7                               & 67.8                               & 65.8                               & \textbf{65.0}                      & {\ul 47.2}                         & 66.6                               \\ \cmidrule{2-12} 
                                                                                                    & CLIP-ft        & 83.6                               & 59.5                               & 46.3                               & 83.6                               & 81.6                               & 78.7                               & 54.2                               & 45.3                               & 33.9                               & 63.0                               \\
                                                                                                    & CLIP-ft-vision & 86.1                               & 56.0                               & 42.2                               & 81.0                               & 79.8                               & 65.1                               & 56.7                               & 42.2                               & 38.7                               & 60.9                               \\
                                                                                                     &  \cellcolor{Gray}Ours           & \cellcolor{Gray}90.8                               & \cellcolor{Gray}\textbf{68.1}                      & \cellcolor{Gray}\textbf{56.0}                      & \cellcolor{Gray}{\ul 89.3}                         & \cellcolor{Gray}\textbf{88.5}                      & \cellcolor{Gray}{\ul 84.3}                         & \cellcolor{Gray}\textbf{70.6}                      & \cellcolor{Gray}{\ul 62.3}                         & \cellcolor{Gray}\textbf{47.7}                      & \cellcolor{Gray}\textbf{73.1}                      \\ \toprule[0.75pt]
\multirow{8}{*}{\begin{tabular}[c]{@{}c@{}}Linear Probe\\ Accuracy (\%)\end{tabular}}               & CLIP           & 95.1                               & 80.5                               & 71.4                               & 93.0                               & {\ul 90.0}                         & \textbf{88.8}                      & 76.6                               & 81.1                               & 76.5                               & 83.7                               \\
                                                                                                    & DeCLIP         & \textbf{96.5}                      & \textbf{84.7}                      & 65.0                               & 94.8                               & 89.2                               & 85.0                               & 75.0                               & 81.6                               & 78.5                               & 83.4                               \\
                                                                                                    & NegCLIP        & 94.3                               & 79.3                               & 71.8                               & 98.7                               & 89.5                               & 85.6                               & 78.6                               & 75.0                               & 81.3                               & 83.8                               \\
                                                                                                    & FILIP          & 95.1                               & 82.4                               & {\ul 77.0}                         & {\ul 99.1}                         & 88.3                               & 83.4                               & {\ul 78.7}                         & 76.8                               & {\ul 88.3}                         & {\ul 85.5}                         \\
                                                                                                    & PyramidCLIP    & {\ul 96.0}                         & 82.5                               & 72.3                               & 96.4                               & 87.8                               & 83.3                               & 77.5                               & {\ul 82.6}                         & 77.3                               & 84.0                               \\ \cmidrule{2-12} 
                                                                                                    & CLIP-ft        & 93.1                               & 76.5                               & 70.7                               & 98.1                               & 88.1                               & 81.7                               & 75.8                               & 58.6                               & 76.3                               & 79.9                               \\
                                                                                                    & CLIP-ft-vision & 93.7                               & 77.9                               & 71.7                               & 98.3                               & 88.6                               & 84.3                               & 76.4                               & 73.9                               & 75.4                               & 82.2                               \\
                                                                                                    & \cellcolor{Gray}Ours           & \cellcolor{Gray}95.6                               & \cellcolor{Gray}{\ul 82.7}                         & \cellcolor{Gray}\textbf{77.2}                      & \cellcolor{Gray}\textbf{99.3}                      & \cellcolor{Gray}\textbf{92.9}                      & \cellcolor{Gray}{\ul 88.1}                         & \cellcolor{Gray}\textbf{79.8}                      & \cellcolor{Gray}\textbf{83.0}                      & \cellcolor{Gray}\textbf{88.9}                      & \cellcolor{Gray}\textbf{87.5}                      \\ \toprule[1pt]
\end{tabular}
}
\label{tab:prediction}
\end{table}

\subsection{Evaluation on Prediction Correctness}
\label{subsec:exp_prediction}
{\bf Zero-shot image classification:}
We compare our model against other state-of-the-art and fine-tuned VLMs on zero-shot image classification tasks.
The results are shown in Tab.~\ref{tab:prediction}.
On the average of nine datasets, our model outperforms the second-best result by 2.6\%.
The results indicate the strong transferability of our model to other vision datasets.

{\bf Linear probe:}
Following the common practice~\cite{radford2021learning, jia2021scaling}, we conduct linear probe experiments on the nine image classification datasets.
As shown in Tab.~\ref{tab:prediction}, our model outperforms the second-best result by 2.0\%.
These results demonstrate the superior vision representations learned by our model.

{\bf Fair comparison with fine-tuned models:}
As shown in Tab.~\ref{tab:prediction}, our model outperforms the best fine-tuned model by 10.1\% and 5.3\% on zero-shot and linear probe results.
This suggests that the proposed Rationale-informed Optimization is essential in improving the model’s performance.

\subsection{Evaluation on Rationale Correctness}
{\bf Rationale localizability:}
We compare our model with state-of-the-art and fine-tuned VLMs.
As shown in Tab.~\ref{tab:cub-part}, our model significantly improves the localization accuracy of rationales by 7.5\% and 6.0\% on {\it CUB-Part}~\cite{saha2022improving} and {\it PartImageNet}~\cite{he2022partimagenet}.
This suggests that even without using explicit region annotations, our method significantly enhances the model's localizability of rationales.

{\bf Rationale disentanglability:}
We compare the rationale disentanglement performance of our model with state-of-the-art and fine-tuned models.
As shown in Tab.~\ref{tab:rationale_disen}, on the average of nine image classification datasets, our model outperforms the second-best result by 36.5\%.
This significant improvement reveals that our model can distinguish between different rationales.

{\bf Fair comparison with fine-tuned models:}
To evaluate whether our model's performance gain can be obtained by solely introducing information from our rationale dataset, we conduct fair comparison experiments with fine-tuned CLIP models.
We compare our model with baseline (CLIP-zs), full model fine-tuning (CLIP-ft), and vision-only fine-tuning (CLIP-ft-vision).
All fine-tuned models use the same CLIP initialization and receive the same language supervision as our model.
As shown in Tabs.~\ref{tab:cub-part}\&~\ref{tab:prediction}, our model outperforms the best fine-tuned model by 9.8\% and 41.1\% on rationale localizability and disentanglability.
This indicates that naive fine-tuning using augmented information without constraints would deteriorate the rationale correctness of the model.

\begin{figure}[t]
  \centering
   \includegraphics[width=1.0\linewidth]{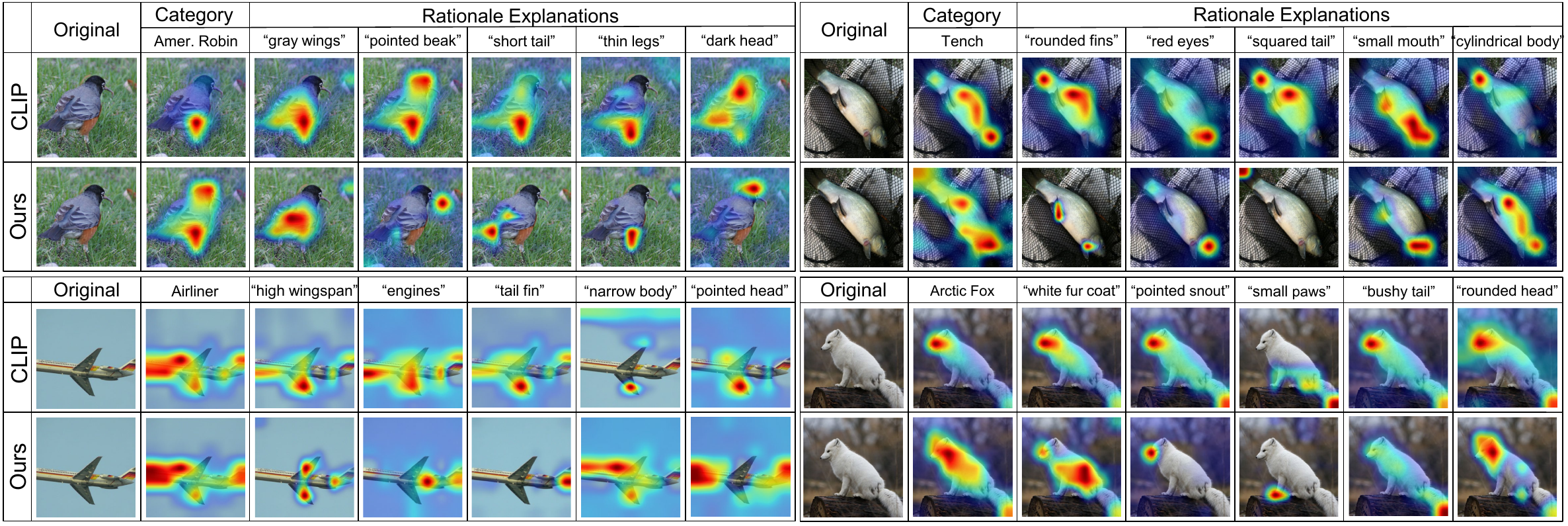}
   \caption{
   Qualitative results of rationale disentanglement and localization.
   The rationales' visual evidence of the CLIP model~\cite{radford2021learning} typically highlights the entire object, lacking precise localization.
   In contrast, our model can correctly localize rationales, thereby enhancing trust in its predictions.
   }
   \label{fig:qualitative}
\end{figure}

{\bf Qualitative results:}
In Fig.~\ref{fig:qualitative}, we show the visualizations of our visual evidence of different rationales.
As shown, the rationales' visual evidence of the CLIP model~\cite{radford2021learning} are entangled and mislocalized.
In contrast, the rationales' visual evidence of our model are visually distinct and correctly localized.

\begin{table}[t]
\centering
\caption{
Comparison of rationale localizability on {\it CUB-Part}~\cite{saha2022improving} and {\it PartImageNet}~\cite{he2022partimagenet}.
As detailed in Sec.~\ref{subsec:evaluation_metrics}, we threshold rationales' explanation heatmaps as segmentation masks and calculate their mIoU ($\uparrow$) with ground truth masks of corresponding object parts.
Our model significantly improves the localization accuracy of fine-grained object parts.
Full table in Appendix~\ref{appen_subsec:full_table}.
}
\resizebox{0.95\columnwidth}{!}{%
\begin{tabular}{lccccccccc}
\toprule[1pt]
\multirow{2}{*}{Models} & \multirow{2}{*}{\begin{tabular}[c]{@{}c@{}}Training\\ Size\end{tabular}} & \multicolumn{7}{c}{CUB-Part}                                                                                  & \multirow{2}{*}{PartImageNet} \\ \cmidrule[0.5pt]{3-9}
                        &                                                                          & Head          & Beak          & Tail          & Wings         & Eyes          & Torso         & Avg.          &                               \\ \toprule[0.75pt]
CLIP                    & 400M                                                                     & {\ul 16.6}    & 3.1           & {\ul 9.9}     & 25.5          & 3.3           & 28.0          & {\ul 14.4}    & {\ul 5.2}                     \\
DeCLIP                  & 88M                                                                      & 6.9           & 2.0           & 5.1           & 16.2          & 1.5           & 18.4          & 8.3           & 3.7                           \\
NegCLIP                 & 400M+COCO                                                                & 15.1          & 3.0           & 7.5           & 26.1          & 2.5           & {\ul 29.4}    & 13.9          & {\ul 5.2}                     \\
FILIP                   & 340M                                                                     & 10.3          & 2.4           & 7.3           & 20.5          & {\ul 3.4}     & 23.7          & 11.2          & 4.0                           \\
PyramidCLIP             & 143M                                                                     & 10.7          & 2.9           & 6.0           & 17.0          & 1.8           & 20.5          & 9.8           & 3.9                           \\ \midrule[0.5pt]
CLIP-ft                 & 400M+IN                                                                  & 13.5          & {\ul 3.3}     & 5.8           & 22.9          & 2.1           & 25.5          & 12.1          & 4.5                           \\
CLIP-ft-vision          & 400M+IN                                                                  & 7.4           & 2.5           & 7.9           & {\ul 26.4}    & 1.6           & 22.0          & 11.3          & 4.4                           \\
\rowcolor{Gray}Ours                    & 400M+IN                                                                  & \textbf{25.3} & \textbf{10.1} & \textbf{12.7} & \textbf{32.6} & \textbf{15.7} & \textbf{35.2} & \textbf{21.9} & \textbf{11.2}                 \\ \toprule[1pt]
\end{tabular}
}
\label{tab:cub-part}
\end{table}

\begin{table}[t]
\centering
\caption
{Comparison of rationale disentanglement. 
We conduct experiments on nine image classification datasets.
Our results are on the average of three trials using different random seeds.
}
\resizebox{0.9\columnwidth}{!}{%
\begin{tabular}{lcccccccccc}
\toprule[1pt]
Models               & C10                              & C100                             & CUB                              & CAL                              & PETS                             & F101                             & SUN                              & CARS                             & DTD                              & AVG                              \\ \toprule[0.75pt]
CLIP                 & 0.249                            & {\ul 0.445}                      & {\ul 0.475}                      & {\ul 0.442}                      & {\ul 0.540}                      & 0.481                            & {\ul 0.519}                      & 0.353                            & 0.287                            & {\ul 0.420}                      \\
DeCLIP               & 0.303                            & 0.297                            & 0.359                            & 0.395                            & 0.388                            & 0.392                            & 0.360                            & 0.332                            & 0.258                            & 0.343                            \\
NegCLIP              & {\ul 0.386}                      & 0.319                            & 0.456                            & 0.401                            & 0.440                            & {\ul 0.491}                      & 0.495                            & {\ul 0.389}                      & 0.261                            & 0.404                            \\
FILIP                & 0.367                            & 0.359                            & 0.267                            & 0.260                            & 0.305                            & 0.384                            & 0.427                            & 0.371                            & {\ul 0.378}                      & 0.346                            \\
PyramidCLIP          & 0.299                            & 0.300                            & 0.428                            & 0.418                            & 0.391                            & 0.318                            & 0.397                            & 0.359                            & 0.283                            & 0.355                            \\ \toprule[0.5pt]
CLIP-ft              & 0.378                            & 0.346                            & 0.469                            & 0.389                            & 0.434                            & 0.374                            & 0.383                            & 0.339                            & 0.251                            & 0.374                            \\
CLIP-ft-vision       & 0.327                            & 0.393                            & 0.433                            & 0.431                            & 0.491                            & 0.475                            & 0.423                            & 0.357                            & 0.274                            & 0.400                            \\
\cellcolor{Gray}Ours & \cellcolor{Gray}{\textbf{0.697}} & \cellcolor{Gray}{\textbf{0.714}} & \cellcolor{Gray}{\textbf{0.831}} & \cellcolor{Gray}{\textbf{0.821}} & \cellcolor{Gray}{\textbf{0.920}} & \cellcolor{Gray}{\textbf{0.823}} & \cellcolor{Gray}{\textbf{0.749}} & \cellcolor{Gray}{\textbf{0.776}} & \cellcolor{Gray}{\textbf{0.734}} & \cellcolor{Gray}{\textbf{0.785}} \\ \toprule[1pt]
\end{tabular}
}
\label{tab:rationale_disen}
\end{table}

\subsection{Ablation Study}

{\bf Ablation on rationale disentanglement:}
The ``w/o disen.'' refers to a variant of our method without rationale disentanglement constraint.
As shown in Tab.~\ref{tab:ablation}, the rationale localizability decreased by 10.4\%, indicating the model might not learn to distinguish between rationales without constraints.

{\bf Ablation on reconstruction:}
The ``w/o recon.'' refers to a variant of our method without reconstruction constraint.
As shown in Tab.~\ref{tab:ablation}, the rationale localizability and prediction accuracy drastically decreased by 13.3\% and 30.2\%.
This reveals that recklessly optimizing the disentanglement between rationale can easily fall into trivial solutions.

{\bf Generalize to different rationale sets:}
According to DCLIP~\cite{menon2022visual}, using the text embeddings of concepts as a bottleneck layer to force the CLIP model~\cite{radford2021learning} to predict based on them can improve prediction accuracy and interpretability.
Specifically, the final prediction will be made by the average embedding similarity between the image and all concepts, namely $\hat{y} = \mathrm{argmax}_y \frac{1}{K}  {\textstyle \sum_{k=1}^{K}} \langle f_\phi(I), f_\theta(c_k^y)\rangle $.
We use the concept set provided in~\cite{menon2022visual} rather than our training rationale dataset.
As shown in Tab.~\ref{tab:random_str}, our model can generalize to an unseen concept set with improved prediction accuracy.

{\bf Ablation using random string:}
WaffleCLIP~\cite{roth2023waffling} shows that random concept strings as bottlenecks can achieve similar performance gains in DCLIP~\cite{menon2022visual}.
We conduct an ablation study using the random strings provided by~\cite{roth2023waffling}.
As shown in Tab.~\ref{tab:random_str}, since our model can distinguish between different rationales, the random strings deteriorate the prediction accuracy of our model.



\subsection{Evaluation on Retrieval Tasks}
\label{subsec:exp_retrieval}
{\bf Zero-shot image-text retrieval:}
We evaluate our model on zero-shot image-text retrieval tasks.
As shown in Tab.~\ref{tab:retrieval}, the improved rationale correctness also benefits retrieval tasks.

\begin{figure}[t]
  \centering
   \includegraphics[width=0.95\linewidth]{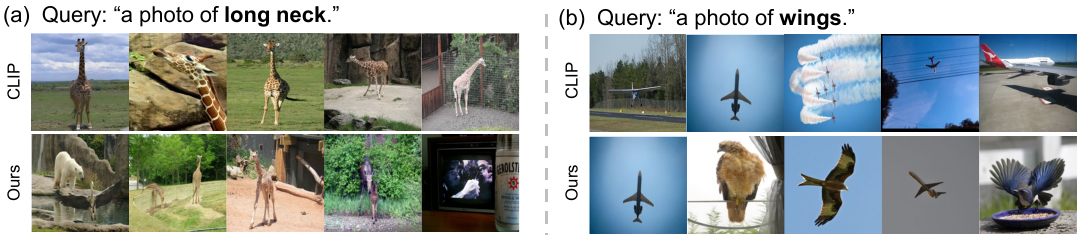}
   \caption{
   Qualitative results of zero-shot text-to-image retrieval on {\it MSCOCO}~\cite{lin2014microsoft}.
   The task is to retrieve the top-5 images with a given rationale presented.
   The CLIP results reveal a significant entangle of rationales with a specific category, such as ``long neck'' with giraffes and ``wings'' with airliners.
   In contrast, our model treats rationales independently from categories, thus offering diverse retrieval results.
   For example, the ``long neck'' found in birds, giraffes, dears, and bottles.
   }
   \label{fig:rationale-retrieval}
\end{figure}

{\bf Rationale-based text-to-image retrieval:}
To better evaluate the rationale correctness of our model, we conduct a novel retrieval task: rationale-based text-to-image retrieval.
The model should retrieve the image with a specified rationale presented.
As shown in Fig.~\ref{fig:rationale-retrieval}, in contrast to the CLIP model~\cite{radford2021learning} that entangles rationales with specific categories, our model precisely understands the semantic meaning of rationales independent to categories.

\begin{table}[t]
    \centering
    \begin{minipage}{0.50\textwidth}
        \centering
        \caption{Comparison of zero-shot image-text retrieval accuracy (\%). Double-correct prediction enhances the model's visual understanding.
        (Note that NegCLIP is trained on {\it MSCOCO}~\cite{lin2014microsoft})}
        \resizebox{0.85\columnwidth}{!}{%
        \begin{tabular}{lcccc}
        \toprule[1pt]
        \multirow{2}{*}{Models} & \multicolumn{2}{c}{MSCOCO} & \multicolumn{2}{c}{Flickr30K} \\ \cmidrule{2-5} 
                                & I2T           & T2I           & I2T          & T2I         \\ \toprule[0.75pt]
        CLIP                    & 32.5          & 28.6          & 64.0         & 60.9        \\
        DeCLIP                  & 32.6          & 22.1          & 59.8         & 46.2        \\
        NegCLIP                 & -             & -             & {\ul 69.3}         & 68.1        \\
        FILIP                   & 33.6          & 36.4          & 52.9         & 53.3        \\
        PyramidCLIP             & {\ul 37.1}          & \textbf{37.6}          & 69.0         & \textbf{69.6}        \\ \midrule[0.5pt]
        CLIP-ft                 & 24.3          & 25.1          & 42.5         & 41.6        \\
        CLIP-ft-vision          & 25.9          & 27.2          & 49.1         & 55.5        \\
        \rowcolor{Gray}Ours                    & \textbf{38.4}          & {\ul 37.1}         & \textbf{69.5}         & {\ul 68.9}        \\ \toprule[1pt]
        \end{tabular}
        }
        \label{tab:retrieval} 
    \end{minipage}
    \hfill
    \begin{minipage}{0.45\textwidth}
        \begin{minipage}{\textwidth}
            \centering
            \captionof{table}{Ablation study on proposed constraints using {\it CUB-Part}~\cite{saha2022improving} and {\it CUB}~\cite{wah2011caltech}.}
            \resizebox{1.0\columnwidth}{!}{%
            \begin{tabular}{lrc}
            \toprule[0.75pt]
            Models          & \multicolumn{1}{c}{mIoU ($\uparrow$)}        & Acc. (\%)                           \\ \toprule[0.5pt]
            Ours w/o disen. & 11.5 $\pm$ \small{1.3}          & 43.3 $\pm$ \small{0.8}          \\
            Ours w/o recon. & 8.6 $\pm$ \small{0.8}          & 25.8 $\pm$ \small{0.7}          \\
            \rowcolor{Gray}Ours (full)     & \textbf{21.9} $\pm$ \small{1.6} & \textbf{56.0} $\pm$ \small{0.5} \\ \toprule[0.75pt]
            \end{tabular}
            }
            \label{tab:ablation}
        \end{minipage}
        
        \begin{minipage}{\textwidth}
            \centering
            \captionof{table}{Comparison of rationale-based prediction accuracy (\%) on {\it ImageNet}~\cite{deng2009imagenet}.}
            \resizebox{0.8\columnwidth}{!}{%
            \begin{tabular}{lccc}
            \toprule[1pt]
            Model        & CLIP & CLIP-ft & Ours          \\ \toprule[0.75pt]
            + concepts   & 63.1 & 67.5    & \textbf{70.5} \\
            + rand. str. & 63.3 & 68.6    & 68.3          \\ \toprule[0.5pt]
            $\Delta$     & {\color{green} +0.2}  & {\color{green} +1.1}     & {\color{red} -2.2}          \\ \bottomrule[1pt]
            \end{tabular}
            }
            \label{tab:random_str}
        \end{minipage}
    \end{minipage}
\end{table}

\section{Related Works}
\label{sec:related_works}
{\bf Vision model explainability.}
A widely adopted branch of explainability methods {\it post hoc} generates heatmaps to identify the image regions most crucial to the model’s predictions, {\it e.g.}, GradCAM~\cite{selvaraju2017grad}, LIME~\cite{ribeiro2016should}, and SHAP~\cite{lundberg2017unified}.
Although useful for revealing the correlations between inputs and outputs, such explanations might be ambiguous, and fail to correspond to high-level concepts that humans easily understand~\cite{rudin2019stop}.
Methods like TCAV~\cite{kim2018interpretability} curate attribute datasets to explain vision models using concepts familiar to humans.
However, such methods can fail when the models do not learn these concepts~\cite{achtibat2022towards}.
Another branch of methods attempts to design specific architecture to {\it intrinsically interpret} model predictions, {\it e.g.}, CBM~\cite{koh2020concept} and ProtoPNet~\cite{chen2019looks}.
However, they cannot guarantee the model learns the semantic meanings of the concepts correctly~\cite{margeloiu2021concept} and yield compromised prediction accuracy~\cite{koh2020concept}.
Different from existing works, our method incorporates explanations to guide the model training, achieving accurate predictions backed by correct rationales.

{\bf Knowledge augmentation for vision-language models.}
Visual models often learn spurious correlations that stem from data biases unrelated to the causal explanation of interest~\cite{arjovsky2019invariant, qiao2020learning}, whereas external knowledge allows models to learn the right features~\cite{chen2019looks, li2021deep}.
Existing attempts for injecting knowledge into the models are often from the language modality.
K-LITE~\cite{shen2022k} enrich the image caption using knowledge from WordNet~\cite{miller1995wordnet} and Wikitionary~\cite{meyer2012wiktionary}.
NegCLIP~\cite{yuksekgonul2022and} and DANCE~\cite{ye2023improving} improve the commonsense understanding of CLIP by generating hard negative captions, the latter uses knowledge from ConceptNet~\cite{speer2017conceptnet}.
StructureCLIP~\cite{huang2024structure} leverages scene-graphs~\cite{krishna2017visual} to incorporate knowledge into text embeddings.
However, our results (Tabs.~\ref{tab:cub-part}\&~\ref{tab:ablation}) reveal that solely augmenting information in the language cannot guarantee the model learning correct features.
In contrast to these works, our method offers supervision signals from both modalities to ensure double-correctness.

{\bf Contrastive vision-language alignment.}
Different from conventional multimodal learning that fuses different modalities~\cite{ma2021smil, ma2022multimodal}, large-scale vison-language pretrained models, such as CLIP~\cite{radford2021learning} and ALIGN~\cite{jia2021scaling}, exhibit promising zero-shot transferability to downstream tasks.
However, their global alignment objective is coarse, which only learns the existence of objects like bag-of-word while ignoring their localizations~\cite{yuksekgonul2022and}.
Recent attempts like PyramidCLIP~\cite{gao2022pyramidclip} and X-VLM~\cite{zeng2022multi} leverage object region annotations to align word phrases with image regions.
DeCLIP~\cite{li2022supervision} and FILIP~\cite{yao2021filip} align text with image regions through self-supervised learning.
However, their supervision is limited to a coarse, object-level granularity.
Different from these works, our method offers fine-grained, concept-level supervisory signals of rationales without expensive manual annotations.

\section{Limitation}
While our study advances the double-correctness of predictions, it is not without limitations. First, the absence of explicit ground truth for rationale localization in large-scale datasets remains a significant challenge. We mitigated this by leveraging a self-supervised rationale disentanglement and localization method, but this approach depends heavily on the quality of the structured rationale ontologies. Second, our methods, though effective, are computationally intensive, which may limit their applicability in resource-constrained scenarios.

\section{Conclusion}
We introduce a new concept of \textit{double-correct predictions} aimed at training vision-language foundation models to make accurate predictions backed by correct rationales, thereby enhancing their safety for real-world deployment. To support this, we establish a solid foundation for the development of double-correct predictions. Specifically, we develop a unique dataset with structured rationales that clearly outline the reasoning processes necessary for visual recognition tasks. Furthermore, we propose a principled rationale-informed optimization method tailored for double-correct prediction. Our comprehensive empirical evaluations demonstrate that our method significantly enhances the double correctness of vision-language model predictions.

\newpage

\section*{Acknowledgments}
This work is supported by the NSF CAREER Award No. 2340074, the NSF SLES Award No. 2416937, the NSF III CORE Award No. 2412675, and the DoD DEPSCoR Award AFOSR FA9550-23-1-0494.
Any opinions, findings and conclusions or recommendations expressed in this material are those of the authors and do not reflect the views of the supporting entities.

\bibliographystyle{unsrt}
\bibliography{main}

\begin{thebibliography}{10}

\bibitem{radford2021learning}
Alec Radford, Jong~Wook Kim, Chris Hallacy, Aditya Ramesh, Gabriel Goh, Sandhini Agarwal, Girish Sastry, Amanda Askell, Pamela Mishkin, Jack Clark, et~al.
\newblock Learning transferable visual models from natural language supervision.
\newblock In {\em International conference on machine learning}, pages 8748--8763. PMLR, 2021.

\bibitem{openai2023gpt4}
Josh Achiam, Steven Adler, Sandhini Agarwal, Lama Ahmad, Ilge Akkaya, Florencia~Leoni Aleman, Diogo Almeida, Janko Altenschmidt, Sam Altman, Shyamal Anadkat, et~al.
\newblock {GPT-4 Technical Report}.
\newblock {\em arXiv preprint arXiv:2303.08774}, 2023.

\bibitem{nori2023capabilities}
Harsha Nori, Nicholas King, Scott~Mayer McKinney, Dean Carignan, and Eric Horvitz.
\newblock Capabilities of gpt-4 on medical challenge problems.
\newblock {\em arXiv preprint arXiv:2303.13375}, 2023.

\bibitem{liang2022effective}
Xiwen Liang, Yangxin Wu, Jianhua Han, Hang Xu, Chunjing Xu, and Xiaodan Liang.
\newblock Effective adaptation in multi-task co-training for unified autonomous driving.
\newblock In {\em Advances in Neural Information Processing Systems}, 2022.

\bibitem{liang2023visual}
Xiwen Liang, Minzhe Niu, Jianhua Han, Hang Xu, Chunjing Xu, and Xiaodan Liang.
\newblock Visual exemplar driven task-prompting for unified perception in autonomous driving.
\newblock In {\em Proceedings of the IEEE/CVF Conference on Computer Vision and Pattern Recognition}, pages 9611--9621, 2023.

\bibitem{teach1981analysis}
Randy~L Teach and Edward~H Shortliffe.
\newblock An analysis of physician attitudes regarding computer-based clinical consultation systems.
\newblock {\em Computers and Biomedical Research}, 14(6):542--558, 1981.

\bibitem{ribeiro2016should}
Marco~Tulio Ribeiro, Sameer Singh, and Carlos Guestrin.
\newblock " why should i trust you?" explaining the predictions of any classifier.
\newblock In {\em Proceedings of the 22nd ACM SIGKDD international conference on knowledge discovery and data mining}, pages 1135--1144, 2016.

\bibitem{rudin2019stop}
Cynthia Rudin.
\newblock Stop explaining black box machine learning models for high stakes decisions and use interpretable models instead.
\newblock {\em Nature Machine Intelligence}, 1(5):206--215, 2019.

\bibitem{menon2022visual}
Sachit Menon and Carl Vondrick.
\newblock Visual classification via description from large language models.
\newblock In {\em The Eleventh International Conference on Learning Representations}, 2022.

\bibitem{yuksekgonul2022post}
Mert Yuksekgonul, Maggie Wang, and James Zou.
\newblock Post-hoc concept bottleneck models.
\newblock In {\em The Eleventh International Conference on Learning Representations}, 2022.

\bibitem{yuksekgonul2022and}
Mert Yuksekgonul, Federico Bianchi, Pratyusha Kalluri, Dan Jurafsky, and James Zou.
\newblock When and why vision-language models behave like bags-of-words, and what to do about it?
\newblock In {\em The Eleventh International Conference on Learning Representations}, 2022.

\bibitem{ye2023improving}
Shuquan Ye, Yujia Xie, Dongdong Chen, Yichong Xu, Lu~Yuan, Chenguang Zhu, and Jing Liao.
\newblock Improving commonsense in vision-language models via knowledge graph riddles.
\newblock In {\em Proceedings of the IEEE/CVF Conference on Computer Vision and Pattern Recognition}, pages 2634--2645, 2023.

\bibitem{li2025deal}
Tang Li, Mengmeng Ma, and Xi~Peng.
\newblock Deal: Disentangle and localize concept-level explanations for vlms.
\newblock In {\em European Conference on Computer Vision}, pages 383--401. Springer, 2025.

\bibitem{margeloiu2021concept}
Andrei Margeloiu, Matthew Ashman, Umang Bhatt, Yanzhi Chen, Mateja Jamnik, and Adrian Weller.
\newblock Do concept bottleneck models learn as intended?
\newblock {\em ICLR Workshop}, 2021.

\bibitem{tong2024eyes}
Shengbang Tong, Zhuang Liu, Yuexiang Zhai, Yi~Ma, Yann LeCun, and Saining Xie.
\newblock Eyes wide shut? exploring the visual shortcomings of multimodal llms.
\newblock In {\em Proceedings of the IEEE/CVF Conference on Computer Vision and Pattern Recognition}, pages 9568--9578, 2024.

\bibitem{fei2006one}
Li~Fei-Fei, Robert Fergus, and Pietro Perona.
\newblock One-shot learning of object categories.
\newblock {\em IEEE transactions on pattern analysis and machine intelligence}, 28(4):594--611, 2006.

\bibitem{wah2011caltech}
Catherine Wah, Steve Branson, Peter Welinder, Pietro Perona, and Serge Belongie.
\newblock The caltech-ucsd birds-200-2011 dataset.
\newblock 2011.

\bibitem{deng2009imagenet}
Jia Deng, Wei Dong, Richard Socher, Li-Jia Li, Kai Li, and Li~Fei-Fei.
\newblock Imagenet: A large-scale hierarchical image database.
\newblock In {\em 2009 IEEE conference on computer vision and pattern recognition}, pages 248--255. Ieee, 2009.

\bibitem{speer2017conceptnet}
Robyn Speer, Joshua Chin, and Catherine Havasi.
\newblock Conceptnet 5.5: An open multilingual graph of general knowledge.
\newblock In {\em Proceedings of the AAAI conference on artificial intelligence}, volume~31, 2017.

\bibitem{miller1995wordnet}
George~A Miller.
\newblock Wordnet: a lexical database for english.
\newblock {\em Communications of the ACM}, 38(11):39--41, 1995.

\bibitem{krishna2017visual}
Ranjay Krishna, Yuke Zhu, Oliver Groth, Justin Johnson, Kenji Hata, Joshua Kravitz, Stephanie Chen, Yannis Kalantidis, Li-Jia Li, David~A Shamma, et~al.
\newblock Visual genome: Connecting language and vision using crowdsourced dense image annotations.
\newblock {\em International journal of computer vision}, 123:32--73, 2017.

\bibitem{tschandl2018ham10000}
Philipp Tschandl, Cliff Rosendahl, and Harald Kittler.
\newblock The ham10000 dataset, a large collection of multi-source dermatoscopic images of common pigmented skin lesions.
\newblock {\em Scientific data}, 5(1):1--9, 2018.

\bibitem{roscher2020explainable}
Ribana Roscher, Bastian Bohn, Marco~F Duarte, and Jochen Garcke.
\newblock Explainable machine learning for scientific insights and discoveries.
\newblock {\em Ieee Access}, 8:42200--42216, 2020.

\bibitem{barsalou2008grounded}
Lawrence~W Barsalou.
\newblock Grounded cognition.
\newblock {\em Annu. Rev. Psychol.}, 59:617--645, 2008.

\bibitem{siskind1994grounding}
Jeffrey~Mark Siskind.
\newblock Grounding language in perception.
\newblock {\em Artificial Intelligence Review}, 8:371--391, 1994.

\bibitem{daneshjou2022skincon}
Roxana Daneshjou, Mert Yuksekgonul, Zhuo~Ran Cai, Roberto Novoa, and James~Y Zou.
\newblock Skincon: A skin disease dataset densely annotated by domain experts for fine-grained debugging and analysis.
\newblock {\em Advances in Neural Information Processing Systems}, 35:18157--18167, 2022.

\bibitem{wang2020self}
Yude Wang, Jie Zhang, Meina Kan, Shiguang Shan, and Xilin Chen.
\newblock Self-supervised equivariant attention mechanism for weakly supervised semantic segmentation.
\newblock In {\em Proceedings of the IEEE/CVF Conference on Computer Vision and Pattern Recognition}, pages 12275--12284, 2020.

\bibitem{wang2017knowledge}
Quan Wang, Zhendong Mao, Bin Wang, and Li~Guo.
\newblock Knowledge graph embedding: A survey of approaches and applications.
\newblock {\em IEEE transactions on knowledge and data engineering}, 29(12):2724--2743, 2017.

\bibitem{ji2021survey}
Shaoxiong Ji, Shirui Pan, Erik Cambria, Pekka Marttinen, and S~Yu Philip.
\newblock A survey on knowledge graphs: Representation, acquisition, and applications.
\newblock {\em IEEE transactions on neural networks and learning systems}, 33(2):494--514, 2021.

\bibitem{bubeck2023sparks}
S{\'e}bastien Bubeck, Varun Chandrasekaran, Ronen Eldan, Johannes Gehrke, Eric Horvitz, Ece Kamar, Peter Lee, Yin~Tat Lee, Yuanzhi Li, Scott Lundberg, et~al.
\newblock Sparks of artificial general intelligence: Early experiments with gpt-4. arxiv.
\newblock {\em arXiv preprint arXiv:2303.12712}, 2023.

\bibitem{liu2023holistic}
Zhengliang Liu, Hanqi Jiang, Tianyang Zhong, Zihao Wu, Chong Ma, Yiwei Li, Xiaowei Yu, Yutong Zhang, Yi~Pan, Peng Shu, et~al.
\newblock Holistic evaluation of gpt-4v for biomedical imaging.
\newblock {\em arXiv preprint arXiv:2312.05256}, 2023.

\bibitem{brown2020language}
Tom Brown, Benjamin Mann, Nick Ryder, Melanie Subbiah, Jared~D Kaplan, Prafulla Dhariwal, Arvind Neelakantan, Pranav Shyam, Girish Sastry, Amanda Askell, et~al.
\newblock Language models are few-shot learners.
\newblock {\em Advances in neural information processing systems}, 33:1877--1901, 2020.

\bibitem{mcintosh2023culturally}
Timothy~R McIntosh, Tong Liu, Teo Susnjak, Paul Watters, Alex Ng, and Malka~N Halgamuge.
\newblock A culturally sensitive test to evaluate nuanced gpt hallucination.
\newblock {\em IEEE Transactions on Artificial Intelligence}, 2023.

\bibitem{chelli2024hallucination}
Mika{\"e}l Chelli, Jules Descamps, Vincent Lavou{\'e}, Christophe Trojani, Michel Azar, Marcel Deckert, Jean-Luc Raynier, Gilles Clowez, Pascal Boileau, and Caroline Ruetsch-Chelli.
\newblock Hallucination rates and reference accuracy of chatgpt and bard for systematic reviews: Comparative analysis.
\newblock {\em Journal of Medical Internet Research}, 26:e53164, 2024.

\bibitem{yang2023language}
Yue Yang, Artemis Panagopoulou, Shenghao Zhou, Daniel Jin, Chris Callison-Burch, and Mark Yatskar.
\newblock Language in a bottle: Language model guided concept bottlenecks for interpretable image classification.
\newblock In {\em Proceedings of the IEEE/CVF Conference on Computer Vision and Pattern Recognition}, pages 19187--19197, 2023.

\bibitem{qin2022medical}
Ziyuan Qin, Hua~Hui Yi, Qicheng Lao, and Kang Li.
\newblock Medical image understanding with pretrained vision language models: A comprehensive study.
\newblock In {\em The Eleventh International Conference on Learning Representations}, 2022.

\bibitem{dosovitskiy2020image}
Alexey Dosovitskiy, Lucas Beyer, Alexander Kolesnikov, Dirk Weissenborn, Xiaohua Zhai, Thomas Unterthiner, Mostafa Dehghani, Matthias Minderer, Georg Heigold, Sylvain Gelly, et~al.
\newblock An image is worth 16x16 words: Transformers for image recognition at scale.
\newblock In {\em International Conference on Learning Representations}, 2020.

\bibitem{zhou2022conditional}
Kaiyang Zhou, Jingkang Yang, Chen~Change Loy, and Ziwei Liu.
\newblock Conditional prompt learning for vision-language models.
\newblock In {\em Proceedings of the IEEE/CVF conference on computer vision and pattern recognition}, pages 16816--16825, 2022.

\bibitem{abnar2020quantifying}
Samira Abnar and Willem Zuidema.
\newblock Quantifying attention flow in transformers.
\newblock {\em arXiv preprint arXiv:2005.00928}, 2020.

\bibitem{chefer2021transformer}
Hila Chefer, Shir Gur, and Lior Wolf.
\newblock Transformer interpretability beyond attention visualization.
\newblock In {\em Proceedings of the IEEE/CVF conference on computer vision and pattern recognition}, pages 782--791, 2021.

\bibitem{selvaraju2017grad}
Ramprasaath~R Selvaraju, Michael Cogswell, Abhishek Das, Ramakrishna Vedantam, Devi Parikh, and Dhruv Batra.
\newblock Grad-cam: Visual explanations from deep networks via gradient-based localization.
\newblock In {\em Proceedings of the IEEE international conference on computer vision}, pages 618--626, 2017.

\bibitem{liu2022rethinking}
Yibing Liu, Haoliang Li, Yangyang Guo, Chenqi Kong, Jing Li, and Shiqi Wang.
\newblock Rethinking attention-model explainability through faithfulness violation test.
\newblock In {\em International Conference on Machine Learning}, pages 13807--13824. PMLR, 2022.

\bibitem{elhage2021mathematical}
Nelson Elhage, Neel Nanda, Catherine Olsson, Tom Henighan, Nicholas Joseph, Ben Mann, Amanda Askell, Yuntao Bai, Anna Chen, Tom Conerly, et~al.
\newblock A mathematical framework for transformer circuits.
\newblock {\em Transformer Circuits Thread}, 1(1):12, 2021.

\bibitem{gandelsman2023interpreting}
Yossi Gandelsman, Alexei~A Efros, and Jacob Steinhardt.
\newblock Interpreting clip's image representation via text-based decomposition.
\newblock In {\em The Twelfth International Conference on Learning Representations}, 2023.

\bibitem{guillaumin2014imagenet}
Matthieu Guillaumin, Daniel K{\"u}ttel, and Vittorio Ferrari.
\newblock Imagenet auto-annotation with segmentation propagation.
\newblock {\em International Journal of Computer Vision}, 110:328--348, 2014.

\bibitem{binder2016layer}
Alexander Binder, Gr{\'e}goire Montavon, Sebastian Lapuschkin, Klaus-Robert M{\"u}ller, and Wojciech Samek.
\newblock Layer-wise relevance propagation for neural networks with local renormalization layers.
\newblock In {\em Artificial Neural Networks and Machine Learning--ICANN 2016: 25th International Conference on Artificial Neural Networks, Barcelona, Spain, September 6-9, 2016, Proceedings, Part II 25}, pages 63--71. Springer, 2016.

\bibitem{li2023dre}
Tang Li, Fengchun Qiao, Mengmeng Ma, and Xi~Peng.
\newblock Are data-driven explanations robust against out-of-distribution data?
\newblock In {\em Proceedings of the IEEE/CVF Conference on Computer Vision and Pattern Recognition}, pages 3821--3831, 2023.

\bibitem{oord2018representation}
Aaron van~den Oord, Yazhe Li, and Oriol Vinyals.
\newblock Representation learning with contrastive predictive coding.
\newblock {\em arXiv preprint arXiv:1807.03748}, 2018.

\bibitem{robey2021model}
Alexander Robey, George~J Pappas, and Hamed Hassani.
\newblock Model-based domain generalization.
\newblock {\em Advances in Neural Information Processing Systems}, 34:20210--20229, 2021.

\bibitem{qiao2023topology}
Fengchun Qiao and Xi~Peng.
\newblock Topology-aware robust optimization for out-of-distribution generalization.
\newblock In {\em Proceedings of the International Conference on Learning Representations (ICLR)}, 2023.

\bibitem{boyd2004convex}
Stephen Boyd, Stephen~P Boyd, and Lieven Vandenberghe.
\newblock {\em Convex optimization}.
\newblock Cambridge university press, 2004.

\bibitem{ma2024beyond}
Mengmeng Ma, Tang Li, and Xi~Peng.
\newblock Beyond the federation: Topology-aware federated learning for generalization to unseen clients.
\newblock In {\em Proceedings of the International Conference on Machine Learning (ICML)}, 2024.

\bibitem{gao2022pyramidclip}
Yuting Gao, Jinfeng Liu, Zihan Xu, Jun Zhang, Ke~Li, Rongrong Ji, and Chunhua Shen.
\newblock Pyramidclip: Hierarchical feature alignment for vision-language model pretraining.
\newblock {\em Advances in neural information processing systems}, 35:35959--35970, 2022.

\bibitem{zeng2022multi}
Yan Zeng, Xinsong Zhang, and Hang Li.
\newblock Multi-grained vision language pre-training: Aligning texts with visual concepts.
\newblock In {\em International Conference on Machine Learning}, pages 25994--26009. PMLR, 2022.

\bibitem{torralba2011unbiased}
Antonio Torralba and Alexei~A Efros.
\newblock Unbiased look at dataset bias.
\newblock In {\em CVPR 2011}, pages 1521--1528. IEEE, 2011.

\bibitem{li2022supervision}
Yangguang Li, Feng Liang, Lichen Zhao, Yufeng Cui, Wanli Ouyang, Jing Shao, Fengwei Yu, and Junjie Yan.
\newblock Supervision exists everywhere: A data efficient contrastive language-image pre-training paradigm.
\newblock In {\em International Conference on Learning Representations}, 2022.

\bibitem{yao2021filip}
Y.~L. et~al.
\newblock Filip: Fine-grained interactive language-image pre-training.
\newblock {\em ICLR}, 2022.

\bibitem{fei2004learning}
Li~Fei-Fei, Rob Fergus, and Pietro Perona.
\newblock Learning generative visual models from few training examples: An incremental bayesian approach tested on 101 object categories.
\newblock In {\em 2004 conference on computer vision and pattern recognition workshop}, pages 178--178. IEEE, 2004.

\bibitem{parkhi2012cats}
Omkar~M Parkhi, Andrea Vedaldi, Andrew Zisserman, and CV~Jawahar.
\newblock Cats and dogs.
\newblock In {\em 2012 IEEE conference on computer vision and pattern recognition}, pages 3498--3505. IEEE, 2012.

\bibitem{bossard2014food}
Lukas Bossard, Matthieu Guillaumin, and Luc Van~Gool.
\newblock Food-101--mining discriminative components with random forests.
\newblock In {\em Computer Vision--ECCV 2014: 13th European Conference, Zurich, Switzerland, September 6-12, 2014, Proceedings, Part VI 13}, pages 446--461. Springer, 2014.

\bibitem{xiao2010sun}
Jianxiong Xiao, James Hays, Krista~A Ehinger, Aude Oliva, and Antonio Torralba.
\newblock Sun database: Large-scale scene recognition from abbey to zoo.
\newblock In {\em 2010 IEEE computer society conference on computer vision and pattern recognition}, pages 3485--3492. IEEE, 2010.

\bibitem{krause20133d}
Jonathan Krause, Michael Stark, Jia Deng, and Li~Fei-Fei.
\newblock 3d object representations for fine-grained categorization.
\newblock In {\em Proceedings of the IEEE international conference on computer vision workshops}, pages 554--561, 2013.

\bibitem{cimpoi2014describing}
Mircea Cimpoi, Subhransu Maji, Iasonas Kokkinos, Sammy Mohamed, and Andrea Vedaldi.
\newblock Describing textures in the wild.
\newblock In {\em Proceedings of the IEEE conference on computer vision and pattern recognition}, pages 3606--3613, 2014.

\bibitem{krizhevsky2009learning}
Alex Krizhevsky, Geoffrey Hinton, et~al.
\newblock Learning multiple layers of features from tiny images.
\newblock 2009.

\bibitem{young2014image}
Peter Young, Alice Lai, Micah Hodosh, and Julia Hockenmaier.
\newblock From image descriptions to visual denotations: New similarity metrics for semantic inference over event descriptions.
\newblock {\em Transactions of the Association for Computational Linguistics}, 2:67--78, 2014.

\bibitem{lin2014microsoft}
Tsung-Yi Lin, Michael Maire, Serge Belongie, James Hays, Pietro Perona, Deva Ramanan, Piotr Doll{\'a}r, and C~Lawrence Zitnick.
\newblock Microsoft coco: Common objects in context.
\newblock In {\em Computer Vision--ECCV 2014: 13th European Conference, Zurich, Switzerland, September 6-12, 2014, Proceedings, Part V 13}, pages 740--755. Springer, 2014.

\bibitem{saha2022improving}
Oindrila Saha, Zezhou Cheng, and Subhransu Maji.
\newblock Improving few-shot part segmentation using coarse supervision.
\newblock In {\em European Conference on Computer Vision}, pages 283--299. Springer, 2022.

\bibitem{he2022partimagenet}
Ju~He, Shuo Yang, Shaokang Yang, Adam Kortylewski, Xiaoding Yuan, Jie-Neng Chen, Shuai Liu, Cheng Yang, Qihang Yu, and Alan Yuille.
\newblock Partimagenet: A large, high-quality dataset of parts.
\newblock In {\em European Conference on Computer Vision}, pages 128--145. Springer, 2022.

\bibitem{loshchilov2018decoupled}
Ilya Loshchilov and Frank Hutter.
\newblock Decoupled weight decay regularization.
\newblock In {\em International Conference on Learning Representations}, 2018.

\bibitem{selva2017grad}
R.~S. et~al.
\newblock Grad-cam: Visual explanations from deep networks via gradient-based localization.
\newblock {\em ICCV}, 2017.

\bibitem{jia2021scaling}
Chao Jia, Yinfei Yang, Ye~Xia, Yi-Ting Chen, Zarana Parekh, Hieu Pham, Quoc Le, Yun-Hsuan Sung, Zhen Li, and Tom Duerig.
\newblock Scaling up visual and vision-language representation learning with noisy text supervision.
\newblock In {\em International conference on machine learning}, pages 4904--4916. PMLR, 2021.

\bibitem{roth2023waffling}
Karsten Roth, Jae~Myung Kim, A~Koepke, Oriol Vinyals, Cordelia Schmid, and Zeynep Akata.
\newblock Waffling around for performance: Visual classification with random words and broad concepts.
\newblock In {\em Proceedings of the IEEE/CVF International Conference on Computer Vision}, pages 15746--15757, 2023.

\bibitem{lundberg2017unified}
Scott~M Lundberg and Su-In Lee.
\newblock A unified approach to interpreting model predictions.
\newblock {\em Advances in neural information processing systems}, 30, 2017.

\bibitem{kim2018interpretability}
Been Kim, Martin Wattenberg, Justin Gilmer, Carrie Cai, James Wexler, Fernanda Viegas, et~al.
\newblock Interpretability beyond feature attribution: Quantitative testing with concept activation vectors (tcav).
\newblock In {\em International conference on machine learning}, pages 2668--2677. PMLR, 2018.

\bibitem{achtibat2022towards}
Reduan Achtibat, Maximilian Dreyer, Ilona Eisenbraun, Sebastian Bosse, Thomas Wiegand, Wojciech Samek, and Sebastian Lapuschkin.
\newblock From" where" to" what": Towards human-understandable explanations through concept relevance propagation.
\newblock {\em arXiv preprint arXiv:2206.03208}, 2022.

\bibitem{koh2020concept}
Pang~Wei Koh, Thao Nguyen, Yew~Siang Tang, Stephen Mussmann, Emma Pierson, Been Kim, and Percy Liang.
\newblock Concept bottleneck models.
\newblock In {\em International conference on machine learning}, pages 5338--5348. PMLR, 2020.

\bibitem{chen2019looks}
Chaofan Chen, Oscar Li, Daniel Tao, Alina Barnett, Cynthia Rudin, and Jonathan~K Su.
\newblock This looks like that: deep learning for interpretable image recognition.
\newblock {\em Advances in neural information processing systems}, 32, 2019.

\bibitem{arjovsky2019invariant}
Martin Arjovsky, L{\'e}on Bottou, Ishaan Gulrajani, and David Lopez-Paz.
\newblock Invariant risk minimization.
\newblock {\em arXiv preprint arXiv:1907.02893}, 2019.

\bibitem{qiao2020learning}
Fengchun Qiao, Long Zhao, and Xi~Peng.
\newblock Learning to learn single domain generalization.
\newblock In {\em Proceedings of the IEEE/CVF Conference on Computer Vision and Pattern Recognition}, pages 12556--12565, 2020.

\bibitem{li2021deep}
Tang Li, Jing Gao, and Xi~Peng.
\newblock Deep learning for spatiotemporal modeling of urbanization.
\newblock {\em Advances in Neural Information Processing Systems Workshops (Best Paper Award)}, 2021.

\bibitem{shen2022k}
S.~S. et~al.
\newblock K-lite: Learning transferable visual models with external knowledge.
\newblock {\em NeurIPS}, 2022.

\bibitem{meyer2012wiktionary}
Christian~M Meyer and Iryna Gurevych.
\newblock {\em Wiktionary: A new rival for expert-built lexicons? Exploring the possibilities of collaborative lexicography}.
\newblock na, 2012.

\bibitem{huang2024structure}
Yufeng Huang, Jiji Tang, Zhuo Chen, Rongsheng Zhang, Xinfeng Zhang, Weijie Chen, Zeng Zhao, Zhou Zhao, Tangjie Lv, Zhipeng Hu, et~al.
\newblock Structure-clip: Towards scene graph knowledge to enhance multi-modal structured representations.
\newblock In {\em Proceedings of the AAAI Conference on Artificial Intelligence}, volume~38, pages 2417--2425, 2024.

\bibitem{ma2021smil}
Mengmeng Ma, Jian Ren, Long Zhao, Sergey Tulyakov, Cathy Wu, and Xi~Peng.
\newblock Smil: Multimodal learning with severely missing modality.
\newblock In {\em Proceedings of the AAAI Conference on Artificial Intelligence}, volume~35, pages 2302--2310, 2021.

\bibitem{ma2022multimodal}
Mengmeng Ma, Jian Ren, Long Zhao, Davide Testuggine, and Xi~Peng.
\newblock Are multimodal transformers robust to missing modality?
\newblock In {\em Proceedings of the IEEE/CVF Conference on Computer Vision and Pattern Recognition}, pages 18177--18186, 2022.

\end{thebibliography}

\newpage
\section*{Appendix}

\appendix

\section{Full Prompts}
\label{appen_subsec:prompts}

\begin{tcolorbox}[breakable, colback=mycolback, colframe=myframecolor,title=Our Prompt to Obtain Structured Rationales]

\begin{verbatim}
American Robin = {
  "nodes": [
    {"id": "American Robin", "label": "American Robin"},
    {"id": "Breast", "label": "Breast"},
    {"id": "Tail", "label": "Tail"},
    {"id": "Beak", "label": "Beak"},
    {"id": "Eyes", "label": "Eyes"},
    {"id": "Red", "label": "Red"},
    {"id": "Gray", "label": "Gray"},
    {"id": "Yellow", "label": "Yellow"},
    {"id": "Round", "label": "Round"},
    {"id": "Long", "label": "Long"}
  ],
  "edges": [
    {"source": "American Robin", "target": "Breast", "relation": "has"},
    {"source": "American Robin", "target": "Tail", "relation": "has"},
    {"source": "American Robin", "target": "Beak", "relation": "has"},
    {"source": "American Robin", "target": "Eyes", "relation": "has"},
    {"source": "Breast", "target": "Red", "relation": "is"},
    {"source": "Tail", "target": "Gray", "relation": "is"},
    {"source": "Beak", "target": "Yellow", "relation": "is"},
    {"source": "Eyes", "target": "Round", "relation": "are"},
    {"source": "Tail", "target": "Long", "relation": "is"}
  ]
}

Airliner = {
  "nodes": [
    {"id": "Airliner", "label": "Airliner"},
    {"id": "Wings", "label": "Wings"},
    {"id": "Tail", "label": "Tail"},
    {"id": "Fuselage", "label": "Fuselage"},
    {"id": "Engines", "label": "Engines"},
    {"id": "Windows", "label": "Windows"},
    {"id": "Logo", "label": "Logo"},
    {"id": "Large", "label": "Large"},
    {"id": "Horizontal stabilizer", "label": "Horizontal stabilizer"},
    {"id": "Cylindrical", "label": "Cylindrical"},
    {"id": "Under wings", "label": "Under wings"},
    {"id": "Rowed", "label": "Rowed"},
    {"id": "Tail fin", "label": "Tail fin"}
  ],
  "edges": [
    {"source": "Airliner", "target": "Wings", "relation": "has"},
    {"source": "Airliner", "target": "Tail", "relation": "has"},
    {"source": "Airliner", "target": "Fuselage", "relation": "has"},
    {"source": "Airliner", "target": "Engines", "relation": "has"},
    {"source": "Airliner", "target": "Windows", "relation": "has"},
    {"source": "Airliner", "target": "Logo", "relation": "has"},
    {"source": "Wings", "target": "Large", "relation": "are"},
    {"source": "Tail", "target": "Horiz. stabilizer", "relation": "has"},
    {"source": "Fuselage", "target": "Cylindrical", "relation": "is"},
    {"source": "Engines", "target": "Under wings", "relation": "are"},
    {"source": "Windows", "target": "Rowed", "relation": "are"},
    {"source": "Tail", "target": "Tail fin", "relation": "has"}
  ]
}

What are useful visual concepts for distinguishing a {category_name} 
in a photo? These features should be visually distinctable and have 
limited overlap with each other. These features should include 
attributes and their relations. For each item, you should be concise 
and precise, and use no more than five words. No ambiguous answers. 
Show your answer using a tree structure in JSON format strictly 
following the examples shown above. Only contains two depths of 
nodes (depth 1: attributes, depth 2: subattributes). No connections 
between node with the same depth. Do not contain a node without an 
edge connected to it. No other explanations, only provide the graph.

\end{verbatim}

\end{tcolorbox}

\section{Machine Evaluation on Full Dataset}
\label{appen_subsec:rationale_eval}

\begin{table}[h]
\centering
\caption{The machine evaluation results on the quality of the full rationale dataset.}
\begin{tabular}{lccc}
\toprule[0.75pt]
Evaluators & Factual Consistency & Comprehensiveness & Visual Disentanglement \\ \midrule[0.5pt]
GPT-4v     & 4.74                & 4.39              & 4.52                   \\
GPT-4o     & 4.89                & 4.59              & 4.61                   \\ \bottomrule[0.75pt]
\end{tabular}
\end{table}

\section{Full Table}
\label{appen_subsec:full_table}

\begin{table}[h]
\centering
\caption{
Evaluation on {\it PartImageNet}~\cite{he2022partimagenet} with ground truth region of parts using ViT-B/32 vision encoder.
We summarize the annotated parts for different categories into 13 common parts.
We apply thresholds to the explanation heatmaps and calculate their mIoU with ground truth masks.
Our model improves the localization accuracy of each part, even though they appear significantly different across categories, such as ``wings'' for birds and airliners.
}
\resizebox{1.0\columnwidth}{!}{%
\begin{tabular}{lcccccccccccccc}
\toprule[1pt]
Model          & Head          & Body          & Foots        & Tail         & Hands         & Fin          & Wings         & Tiers         & Mirror  & Seat         & Seal          & Engine       & Mouth         & Avg.          \\ \toprule[0.75pt]
CLIP           & {\ul 8.4}     & {\ul 9.6}     & 3.9          & 2.5          & 4.7           & {\ul 3.5}    & 5.5           & {\ul 4.2}     & {\ul 0.9}    & {\ul 2.5}    & 11.1          & {\ul 3.8}    & {\ul 7.6}     & {\ul 5.2}     \\
DeCLIP         & 5.9           & 6.5           & 3.2          & 2.4          & 3.7           & 1.8          & 5.2           & 3.0           & 0.5          & 1.7          & 5.4           & 2.7          & 5.9           & 3.7           \\
NegCLIP        & 8.3           & 8.1           & {\ul 4.7}    & 2.1          & 5.2           & 3.7          & 5.7           & 4.1           & 0.7          & 2.0          & {\ul 12}      & 3.7          & 6.9           & {\ul 5.2}     \\
FILIP          & 5.4           & 6.9           & 3.2          & {\ul 2.6}    & 3.9           & 3.0          & {\ul 5.8}     & 3.4           & 0.5          & 1.8          & 6.9           & 3.1          & 4.9           & 4.0           \\
PyramidCLIP    & 4.9           & 7.4           & 3.3          & 1.7          & {\ul 5.7}     & 2.8          & 5.2           & 3.8           & 0.8          & 1.6          & 7.7           & 1.8          & 4.0           & 3.9           \\ \toprule[0.5pt]
CLIP-ft        & 6.9           & 7.3           & 3.5          & 1.9          & 4.9           & 2.6          & 5.5           & {\ul 4.2}     & 0.5          & 1.9          & 9.7           & 2.6          & 6.5           & 4.5           \\
CLIP-ft-vision & 6.4           & 6.9           & 3.4          & 2.1          & 4.4           & 2.5          & 5.4           & 4.1           & 0.7          & 2.1          & 9.3           & 2.9          & 6.4           & 4.4           \\
Ours           & \textbf{16.7} & \textbf{21.1} & \textbf{8.9} & \textbf{7.2} & \textbf{10.4} & \textbf{7.9} & \textbf{11.5} & \textbf{10.1} & \textbf{3.2} & \textbf{5.5} & \textbf{26.3} & \textbf{4.7} & \textbf{12.3} & \textbf{11.2} \\ \bottomrule[1pt]
\end{tabular}
}
\label{tab:part-imagenet}
\end{table}

\section{Implementation Details}
\label{appen_subsec:details}

\begin{table}[h]
\centering
\caption{Datasets for classification task.}
\begin{tabular}{ccccc}
\toprule[0.75pt]
Dataset              & Abbreviation & Classes & Train Size & \multicolumn{1}{l}{Test Size} \\ \toprule[0.5pt]
CIFAR-10             & C10          & 10      & 50,000     & 10,000                        \\
CIFAR-100            & C100         & 100     & 50,000     & 10,000                        \\
Describable Textures & DTD          & 47      & 3,760      & 1,880                         \\
Stanford Cars        & CARS         & 196     & 8,144      & 8,041                         \\
Food-101             & F101         & 101     & 75,750     & 25,250                        \\
Oxford-IIIT Pets     & PETS         & 37      & 3,680      & 3,669                         \\
SUN397               & SUN          & 397     & 19,850     & 19,850                        \\
Caltech-101          & CAL          & 102     & 3,060      & 6,085                         \\
CUB-200-2011         & CUB          & 200     & 5,994      & 5,794                         \\ \bottomrule[0.75pt]
\end{tabular}
\end{table}

\newpage
\section*{NeurIPS Paper Checklist}

\begin{enumerate}

\item {\bf Claims}
    \item[] Question: Do the main claims made in the abstract and introduction accurately reflect the paper's contributions and scope?
    \item[] Answer: \answerYes{} 
    \item[] Justification: The abstract and introduction clearly outline the main claims of the paper, delineating both the theoretical and experimental contributions, which are supported by the results presented. 
    \item[] Guidelines:
    \begin{itemize}
        \item The answer NA means that the abstract and introduction do not include the claims made in the paper.
        \item The abstract and/or introduction should clearly state the claims made, including the contributions made in the paper and important assumptions and limitations. A No or NA answer to this question will not be perceived well by the reviewers. 
        \item The claims made should match theoretical and experimental results, and reflect how much the results can be expected to generalize to other settings. 
        \item It is fine to include aspirational goals as motivation as long as it is clear that these goals are not attained by the paper. 
    \end{itemize}

\item {\bf Limitations}
    \item[] Question: Does the paper discuss the limitations of the work performed by the authors?
    \item[] Answer: \answerYes{} 
    \item[] Justification: The paper comprehensively discusses the limitations of the proposed methods, including robustness against violations of underlying assumptions and scalability concerns.
    \item[] Guidelines:
    \begin{itemize}
        \item The answer NA means that the paper has no limitation while the answer No means that the paper has limitations, but those are not discussed in the paper. 
        \item The authors are encouraged to create a separate "Limitations" section in their paper.
        \item The paper should point out any strong assumptions and how robust the results are to violations of these assumptions (e.g., independence assumptions, noiseless settings, model well-specification, asymptotic approximations only holding locally). The authors should reflect on how these assumptions might be violated in practice and what the implications would be.
        \item The authors should reflect on the scope of the claims made, e.g., if the approach was only tested on a few datasets or with a few runs. In general, empirical results often depend on implicit assumptions, which should be articulated.
        \item The authors should reflect on the factors that influence the performance of the approach. For example, a facial recognition algorithm may perform poorly when image resolution is low or images are taken in low lighting. Or a speech-to-text system might not be used reliably to provide closed captions for online lectures because it fails to handle technical jargon.
        \item The authors should discuss the computational efficiency of the proposed algorithms and how they scale with dataset size.
        \item If applicable, the authors should discuss possible limitations of their approach to address problems of privacy and fairness.
        \item While the authors might fear that complete honesty about limitations might be used by reviewers as grounds for rejection, a worse outcome might be that reviewers discover limitations that aren't acknowledged in the paper. The authors should use their best judgment and recognize that individual actions in favor of transparency play an important role in developing norms that preserve the integrity of the community. Reviewers will be specifically instructed to not penalize honesty concerning limitations.
    \end{itemize}

\item {\bf Theory Assumptions and Proofs}
    \item[] Question: For each theoretical result, does the paper provide the full set of assumptions and a complete (and correct) proof?
    \item[] Answer: \answerYes{} 
    \item[] Justification: The paper provides a detailed presentation of full assumptions and definitions, as presented in Secs.~\ref{sec:problem}\&~\ref{sec:method}. Each equation and its definitions are clearly numbered and cross-referenced.
    \item[] Guidelines:
    \begin{itemize}
        \item The answer NA means that the paper does not include theoretical results. 
        \item All the theorems, formulas, and proofs in the paper should be numbered and cross-referenced.
        \item All assumptions should be clearly stated or referenced in the statement of any theorems.
        \item The proofs can either appear in the main paper or the supplemental material, but if they appear in the supplemental material, the authors are encouraged to provide a short proof sketch to provide intuition. 
        \item Inversely, any informal proof provided in the core of the paper should be complemented by formal proofs provided in appendix or supplemental material.
        \item Theorems and Lemmas that the proof relies upon should be properly referenced. 
    \end{itemize}

    \item {\bf Experimental Result Reproducibility}
    \item[] Question: Does the paper fully disclose all the information needed to reproduce the main experimental results of the paper to the extent that it affects the main claims and/or conclusions of the paper (regardless of whether the code and data are provided or not)?
    \item[] Answer: \answerYes{} 
    \item[] Justification: The paper fully discloses all necessary details for reproducing the main experimental results, including comprehensive descriptions of the methodologies, experimental setups, and parameter settings. In addition, the code and specific datasets are provided as well.
    \item[] Guidelines:
    \begin{itemize}
        \item The answer NA means that the paper does not include experiments.
        \item If the paper includes experiments, a No answer to this question will not be perceived well by the reviewers: Making the paper reproducible is important, regardless of whether the code and data are provided or not.
        \item If the contribution is a dataset and/or model, the authors should describe the steps taken to make their results reproducible or verifiable. 
        \item Depending on the contribution, reproducibility can be accomplished in various ways. For example, if the contribution is a novel architecture, describing the architecture fully might suffice, or if the contribution is a specific model and empirical evaluation, it may be necessary to either make it possible for others to replicate the model with the same dataset, or provide access to the model. In general. releasing code and data is often one good way to accomplish this, but reproducibility can also be provided via detailed instructions for how to replicate the results, access to a hosted model (e.g., in the case of a large language model), releasing of a model checkpoint, or other means that are appropriate to the research performed.
        \item While NeurIPS does not require releasing code, the conference does require all submissions to provide some reasonable avenue for reproducibility, which may depend on the nature of the contribution. For example
        \begin{enumerate}
            \item If the contribution is primarily a new algorithm, the paper should make it clear how to reproduce that algorithm.
            \item If the contribution is primarily a new model architecture, the paper should describe the architecture clearly and fully.
            \item If the contribution is a new model (e.g., a large language model), then there should either be a way to access this model for reproducing the results or a way to reproduce the model (e.g., with an open-source dataset or instructions for how to construct the dataset).
            \item We recognize that reproducibility may be tricky in some cases, in which case authors are welcome to describe the particular way they provide for reproducibility. In the case of closed-source models, it may be that access to the model is limited in some way (e.g., to registered users), but it should be possible for other researchers to have some path to reproducing or verifying the results.
        \end{enumerate}
    \end{itemize}

\item {\bf Open access to data and code}
    \item[] Question: Does the paper provide open access to the data and code, with sufficient instructions to faithfully reproduce the main experimental results, as described in supplemental material?
    \item[] Answer: \answerYes{} 
    \item[] Justification: The paper provides an Anonymous GitHub link with open access to both the data and code used in the experiments, complete with detailed instructions in the supplemental material that enable faithful reproduction of the main experimental results. This includes exact commands, necessary environment details, and scripts for preprocessing data, ensuring that other researchers can replicate the study's findings without ambiguity.
    \item[] Guidelines:
    \begin{itemize}
        \item The answer NA means that paper does not include experiments requiring code.
        \item Please see the NeurIPS code and data submission guidelines (\url{https://nips.cc/public/guides/CodeSubmissionPolicy}) for more details.
        \item While we encourage the release of code and data, we understand that this might not be possible, so “No” is an acceptable answer. Papers cannot be rejected simply for not including code, unless this is central to the contribution (e.g., for a new open-source benchmark).
        \item The instructions should contain the exact command and environment needed to run to reproduce the results. See the NeurIPS code and data submission guidelines (\url{https://nips.cc/public/guides/CodeSubmissionPolicy}) for more details.
        \item The authors should provide instructions on data access and preparation, including how to access the raw data, preprocessed data, intermediate data, and generated data, etc.
        \item The authors should provide scripts to reproduce all experimental results for the new proposed method and baselines. If only a subset of experiments are reproducible, they should state which ones are omitted from the script and why.
        \item At submission time, to preserve anonymity, the authors should release anonymized versions (if applicable).
        \item Providing as much information as possible in supplemental material (appended to the paper) is recommended, but including URLs to data and code is permitted.
    \end{itemize}

\item {\bf Experimental Setting/Details}
    \item[] Question: Does the paper specify all the training and test details (e.g., data splits, hyperparameters, how they were chosen, type of optimizer, etc.) necessary to understand the results?
    \item[] Answer: \answerYes{} 
    \item[] Justification: The paper details all aspects of the experimental settings, including data splits, hyperparameter selection processes, and the types of optimizers used. 
    \item[] Guidelines:
    \begin{itemize}
        \item The answer NA means that the paper does not include experiments.
        \item The experimental setting should be presented in the core of the paper to a level of detail that is necessary to appreciate the results and make sense of them.
        \item The full details can be provided either with the code, in appendix, or as supplemental material.
    \end{itemize}

\item {\bf Experiment Statistical Significance}
    \item[] Question: Does the paper report error bars suitably and correctly defined or other appropriate information about the statistical significance of the experiments?
    \item[] Answer: \answerYes{} 
    \item[] Justification: The paper provides statistical measures such as error bars and confidence intervals for all major experimental results, such as Tab.~\ref{tab:ablation}. These measures are correctly defined, and the paper details the variability factors they capture, including train/test splits and initialization randomness. 
    \item[] Guidelines:
    \begin{itemize}
        \item The answer NA means that the paper does not include experiments.
        \item The authors should answer "Yes" if the results are accompanied by error bars, confidence intervals, or statistical significance tests, at least for the experiments that support the main claims of the paper.
        \item The factors of variability that the error bars are capturing should be clearly stated (for example, train/test split, initialization, random drawing of some parameter, or overall run with given experimental conditions).
        \item The method for calculating the error bars should be explained (closed form formula, call to a library function, bootstrap, etc.)
        \item The assumptions made should be given (e.g., Normally distributed errors).
        \item It should be clear whether the error bar is the standard deviation or the standard error of the mean.
        \item It is OK to report 1-sigma error bars, but one should state it. The authors should preferably report a 2-sigma error bar than state that they have a 96\% CI, if the hypothesis of Normality of errors is not verified.
        \item For asymmetric distributions, the authors should be careful not to show in tables or figures symmetric error bars that would yield results that are out of range (e.g. negative error rates).
        \item If error bars are reported in tables or plots, The authors should explain in the text how they were calculated and reference the corresponding figures or tables in the text.
    \end{itemize}

\item {\bf Experiments Compute Resources}
    \item[] Question: For each experiment, does the paper provide sufficient information on the computer resources (type of compute workers, memory, time of execution) needed to reproduce the experiments?
    \item[] Answer: \answerYes{} 
    \item[] Justification: The paper adequately details the computational resources required for each experiment, including the types of compute workers (CPU or GPU), memory specifications, and execution times.
    \item[] Guidelines:
    \begin{itemize}
        \item The answer NA means that the paper does not include experiments.
        \item The paper should indicate the type of compute workers CPU or GPU, internal cluster, or cloud provider, including relevant memory and storage.
        \item The paper should provide the amount of compute required for each of the individual experimental runs as well as estimate the total compute. 
        \item The paper should disclose whether the full research project required more compute than the experiments reported in the paper (e.g., preliminary or failed experiments that didn't make it into the paper). 
    \end{itemize}
    
\item {\bf Code Of Ethics}
    \item[] Question: Does the research conducted in the paper conform, in every respect, with the NeurIPS Code of Ethics \url{https://neurips.cc/public/EthicsGuidelines}?
    \item[] Answer: \answerYes{} 
    \item[] Justification: The research presented in the paper adheres fully to the NeurIPS Code of Ethics, ensuring ethical considerations are addressed and complied with throughout the study.
    \item[] Guidelines:
    \begin{itemize}
        \item The answer NA means that the authors have not reviewed the NeurIPS Code of Ethics.
        \item If the authors answer No, they should explain the special circumstances that require a deviation from the Code of Ethics.
        \item The authors should make sure to preserve anonymity (e.g., if there is a special consideration due to laws or regulations in their jurisdiction).
    \end{itemize}

\item {\bf Broader Impacts}
    \item[] Question: Does the paper discuss both potential positive societal impacts and negative societal impacts of the work performed?
    \item[] Answer: \answerYes{} 
    \item[] Justification: The paper effectively discusses both the potential positive and negative societal impacts of the research conducted. It acknowledges the benefits of the proposed technology in enhancing data-driven decision-making processes while also addressing possible negative implications, such as unsafe predictions.
    \item[] Guidelines:
    \begin{itemize}
        \item The answer NA means that there is no societal impact of the work performed.
        \item If the authors answer NA or No, they should explain why their work has no societal impact or why the paper does not address societal impact.
        \item Examples of negative societal impacts include potential malicious or unintended uses (e.g., disinformation, generating fake profiles, surveillance), fairness considerations (e.g., deployment of technologies that could make decisions that unfairly impact specific groups), privacy considerations, and security considerations.
        \item The conference expects that many papers will be foundational research and not tied to particular applications, let alone deployments. However, if there is a direct path to any negative applications, the authors should point it out. For example, it is legitimate to point out that an improvement in the quality of generative models could be used to generate deepfakes for disinformation. On the other hand, it is not needed to point out that a generic algorithm for optimizing neural networks could enable people to train models that generate Deepfakes faster.
        \item The authors should consider possible harms that could arise when the technology is being used as intended and functioning correctly, harms that could arise when the technology is being used as intended but gives incorrect results, and harms following from (intentional or unintentional) misuse of the technology.
        \item If there are negative societal impacts, the authors could also discuss possible mitigation strategies (e.g., gated release of models, providing defenses in addition to attacks, mechanisms for monitoring misuse, mechanisms to monitor how a system learns from feedback over time, improving the efficiency and accessibility of ML).
    \end{itemize}
    
\item {\bf Safeguards}
    \item[] Question: Does the paper describe safeguards that have been put in place for responsible release of data or models that have a high risk for misuse (e.g., pretrained language models, image generators, or scraped datasets)?
    \item[] Answer: \answerYes{} 
    \item[] Justification: The paper outlines comprehensive safeguards for the responsible release of data and models, particularly those with potential for high misuse.
    \item[] Guidelines:
    \begin{itemize}
        \item The answer NA means that the paper poses no such risks.
        \item Released models that have a high risk for misuse or dual-use should be released with necessary safeguards to allow for controlled use of the model, for example by requiring that users adhere to usage guidelines or restrictions to access the model or implementing safety filters. 
        \item Datasets that have been scraped from the Internet could pose safety risks. The authors should describe how they avoided releasing unsafe images.
        \item We recognize that providing effective safeguards is challenging, and many papers do not require this, but we encourage authors to take this into account and make a best faith effort.
    \end{itemize}

\item {\bf Licenses for existing assets}
    \item[] Question: Are the creators or original owners of assets (e.g., code, data, models), used in the paper, properly credited and are the license and terms of use explicitly mentioned and properly respected?
    \item[] Answer: \answerYes{} 
    \item[] Justification: The paper properly credits the creators and original owners of all assets used, including datasets, models, and code. Each asset is clearly cited, with references to the original sources and explicit mention of licenses and terms of use.
    \item[] Guidelines:
    \begin{itemize}
        \item The answer NA means that the paper does not use existing assets.
        \item The authors should cite the original paper that produced the code package or dataset.
        \item The authors should state which version of the asset is used and, if possible, include a URL.
        \item The name of the license (e.g., CC-BY 4.0) should be included for each asset.
        \item For scraped data from a particular source (e.g., website), the copyright and terms of service of that source should be provided.
        \item If assets are released, the license, copyright information, and terms of use in the package should be provided. For popular datasets, \url{paperswithcode.com/datasets} has curated licenses for some datasets. Their licensing guide can help determine the license of a dataset.
        \item For existing datasets that are re-packaged, both the original license and the license of the derived asset (if it has changed) should be provided.
        \item If this information is not available online, the authors are encouraged to reach out to the asset's creators.
    \end{itemize}

\item {\bf New Assets}
    \item[] Question: Are new assets introduced in the paper well documented and is the documentation provided alongside the assets?
    \item[] Answer: \answerYes{} 
    \item[] Justification: The new assets introduced in the paper, including datasets and models, are well documented with comprehensive details provided in structured templates.
    \item[] Guidelines:
    \begin{itemize}
        \item The answer NA means that the paper does not release new assets.
        \item Researchers should communicate the details of the dataset/code/model as part of their submissions via structured templates. This includes details about training, license, limitations, etc. 
        \item The paper should discuss whether and how consent was obtained from people whose asset is used.
        \item At submission time, remember to anonymize your assets (if applicable). You can either create an anonymized URL or include an anonymized zip file.
    \end{itemize}

\item {\bf Crowdsourcing and Research with Human Subjects}
    \item[] Question: For crowdsourcing experiments and research with human subjects, does the paper include the full text of instructions given to participants and screenshots, if applicable, as well as details about compensation (if any)? 
    \item[] Answer: \answerNA{} 
    \item[] Justification: The paper does not involve any experiments or research activities that include crowdsourcing or direct interactions with human subjects. 
    \item[] Guidelines:
    \begin{itemize}
        \item The answer NA means that the paper does not involve crowdsourcing nor research with human subjects.
        \item Including this information in the supplemental material is fine, but if the main contribution of the paper involves human subjects, then as much detail as possible should be included in the main paper. 
        \item According to the NeurIPS Code of Ethics, workers involved in data collection, curation, or other labor should be paid at least the minimum wage in the country of the data collector. 
    \end{itemize}

\item {\bf Institutional Review Board (IRB) Approvals or Equivalent for Research with Human Subjects}
    \item[] Question: Does the paper describe potential risks incurred by study participants, whether such risks were disclosed to the subjects, and whether Institutional Review Board (IRB) approvals (or an equivalent approval/review based on the requirements of your country or institution) were obtained?
    \item[] Answer: \answerNA{} 
    \item[] Justification:  The paper does not involve any experiments or research activities that include crowdsourcing or direct interactions with human subjects.
    \item[] Guidelines:
    \begin{itemize}
        \item The answer NA means that the paper does not involve crowdsourcing nor research with human subjects.
        \item Depending on the country in which research is conducted, IRB approval (or equivalent) may be required for any human subjects research. If you obtained IRB approval, you should clearly state this in the paper. 
        \item We recognize that the procedures for this may vary significantly between institutions and locations, and we expect authors to adhere to the NeurIPS Code of Ethics and the guidelines for their institution. 
        \item For initial submissions, do not include any information that would break anonymity (if applicable), such as the institution conducting the review.
    \end{itemize}

\end{enumerate}

\end{document}